\documentclass{article}

     \PassOptionsToPackage{numbers, compress}{natbib}

\usepackage[main, final]{neurips_2026}

\usepackage[utf8]{inputenc} 
\usepackage[T1]{fontenc}    
\usepackage{hyperref}       
\usepackage{url}            
\usepackage{booktabs}       
\usepackage{amsfonts}       
\usepackage{nicefrac}       
\usepackage{microtype}      
\usepackage{xcolor}         
\usepackage{amsmath}
\usepackage{graphicx}
\usepackage{subcaption}
\usepackage{hyperref}
\usepackage[multiple]{footmisc}
\usepackage{url}
\usepackage{multirow}
\usepackage{makecell}
\usepackage{booktabs}
\usepackage[table]{xcolor}
\usepackage{colortbl}
\usepackage{svg}
\usepackage{algorithm}
\usepackage{algpseudocode}
\usepackage{tcolorbox}
\usepackage{enumitem}
\usepackage{amssymb}
\usepackage{bbding}
\usepackage{pifont}
\usepackage{wasysym}
\usepackage{utfsym}
\usepackage{fontawesome}
\usepackage{amsthm}
\usepackage{tabularx}
\usepackage{wrapfig}
\usepackage{lipsum}
\let\savedCross\Cross      
\let\Cross\relax          
\usepackage{marvosym}
\let\Cross\savedCross

\newtheorem{theorem}{Theorem}[section]
\newtheorem{proposition}[theorem]{Proposition}

\theoremstyle{definition}
\newtheorem{definition}[theorem]{Definition}

\theoremstyle{remark}
\newtheorem{remark}[theorem]{Remark}

\newcommand{\ie}[0]{{{\textit{i.e.}}}}

\definecolor{colorFst}{HTML}{bde6cd}       
\definecolor{colorSnd}{HTML}{e4eebc}       
\definecolor{colorTrd}{HTML}{fff8c5}       

\definecolor{navyblue}{HTML}{0071BC}
\definecolor{hotpink}{HTML}{FF0080}
\definecolor{oai-white}{HTML}{FFFFFF}
\definecolor{oai-black}{HTML}{000000}
\definecolor{oai-red}{HTML}{FF4500}
\definecolor{oai-green}{HTML}{51DA4C}
\definecolor{oai-blue}{HTML}{0000FF}
\definecolor{oai-yellow}{HTML}{FFF639}
\definecolor{oai-magenta}{HTML}{FF45FF}
\definecolor{oai-cyan}{HTML}{00FFFF}
\definecolor{oai-orange}{HTML}{FE7600}
\definecolor{oai-violet}{HTML}{8A2BE2}
\definecolor{oai-brown}{HTML}{A0522D}
\definecolor{oai-green-050}{HTML}{F4FFF4}
\definecolor{oai-green-100}{HTML}{E9FFE8}
\definecolor{oai-green-200}{HTML}{D9FFD8}
\definecolor{oai-green-300}{HTML}{C9FFC7}
\definecolor{oai-green-400}{HTML}{A6FFA3}
\definecolor{oai-green-500}{HTML}{7CF178}
\definecolor{oai-green-600}{HTML}{51DA4C}
\definecolor{oai-green-700}{HTML}{3FA93B}
\definecolor{oai-green-800}{HTML}{2D712A}
\definecolor{oai-green-900}{HTML}{193718}
\definecolor{oai-gray-000}{HTML}{FFFFFF}
\definecolor{oai-gray-100}{HTML}{FAFAFA}
\definecolor{oai-gray-200}{HTML}{F5F5F5}
\definecolor{oai-gray-300}{HTML}{E5E5E5}
\definecolor{oai-gray-400}{HTML}{FFB7A4}
\definecolor{oai-gray-500}{HTML}{CDCDCD}
\definecolor{oai-gray-600}{HTML}{A8A8A8}
\definecolor{oai-gray-700}{HTML}{747474}
\definecolor{oai-gray-800}{HTML}{393939}
\definecolor{oai-gray-900}{HTML}{000000}

\definecolor{mygreen}{HTML}{3cb44b}

\definecolor{myred}{HTML}{E33222}
\definecolor{gr}{RGB}{0, 146, 0}

\title{Lie Routing Vision Language Models for View-Robust 3D Understanding}

\author{%
  Jingjun Yi$^{1,2}\dag$\thanks{Jingjun Yi is currently affinitied with University of Amsterdam. This work was conducted when he was affinitied with Westlake University.}, Chen Hu$^{1}\dag$, Qi Bi$^{2}\textsuperscript{\Letter}$, Hao Zheng$^{3}\textsuperscript{\Letter}$, Huimin Huang$^{3}$, Yongkang Li$^{2}$,\\
  \textbf{Wei Ji}$^{3}$, 
  \textbf{Yawen Huang}$^{3}$, \textbf{Xian Wu}$^{3}$, \textbf{Yefeng Zheng}$^{1}\textsuperscript{\Letter}$ \\
  $\dag$: equal contribution; \quad $\textsuperscript{\Letter}$: correspondence \\
  \textsuperscript{\rm 1}Westlake University, China, 
  \textsuperscript{\rm 2}University of Amsterdam, the Netherland, \\
  \textsuperscript{\rm 3}Tencent, China \\
  \texttt{q.bi@ieee.org, howzheng@tencent.com} \\ \texttt{zhengyefeng@westlake.edu.cn} \\
}

\begin{document}

\maketitle

\begin{abstract}
Multimodal Large Language Models (MLLMs) have made substantial progress in 3D scene understanding from multi-view RGB observations.
Yet a fixed object and query can elicit inconsistent target-related responses across camera viewpoints, a phenomenon we characterize as viewpoint-induced semantic drift.
We introduce LieVLM, a Lie routing framework for view-robust 3D understanding.
Motivated by the rotational component of viewpoint change, Lie routing adapts visual representations by applying local $\mathrm{SO}(3)$ transformations to projected features within learned three-dimensional subspaces.
Geometry tokens extracted by VGGT predict local rotation parameters and channel-wise routing coefficients.
A closed-form Pad\'e--Cayley map transforms projected visual features, whose identity-relative corrections are projected back and gated into the original visual and geometry features.
This design retains the pretrained semantic and direct spatial context while learning view-dependent feature adjustments.
Experiments on 3D captioning, grounding, detection, and spatial reasoning show consistent improvements over additive fusion across training scales and backbones.
Controlled ablations examine the roles of geometric conditioning and transformation structure.
Viewpoint analyses further reveal systematic variation in the predicted routing states alongside more localized target responses and stronger task performance across viewpoint changes.
\end{abstract}

\section{Introduction}
\label{sec:intro}

3D Multimodal Large Language Models (MLLMs) extend visual-language reasoning to scene-level tasks such as object captioning, language-guided localization, and spatial question answering.
Early approaches primarily connect language models to explicit 3D representations, including point clouds and object-level scene features~\citep{xu2024pointllm,chen2024ll3da,huang2024chat}.
These representations provide direct spatial structure, while their reliance on reconstructed scene inputs motivates complementary approaches that learn 3D understanding from RGB observations.

Recent 3D MLLMs derive geometry directly from multi-view RGB observations, enabling 3D understanding without requiring preconstructed 3D scene models.
Pretrained geometry encoders such as VGGT~\citep{wang2025vggt} extract scene structure and cross-view correspondences, and recent methods inject these features into MLLM visual representations for 3D scene understanding and spatial reasoning~\citep{zheng2025learning,huang20253drs,zhang2026generalization}. This learned geometry injection paradigm preserves the scalability of video-based MLLMs while providing strong spatial context. Despite these advances, geometry is primarily used to augment visual representations, while how it should drive viewpoint-adaptive changes in visual semantics remains underexplored.

\begin{figure}[!t]
    \centering
    \includegraphics[width=1\linewidth]{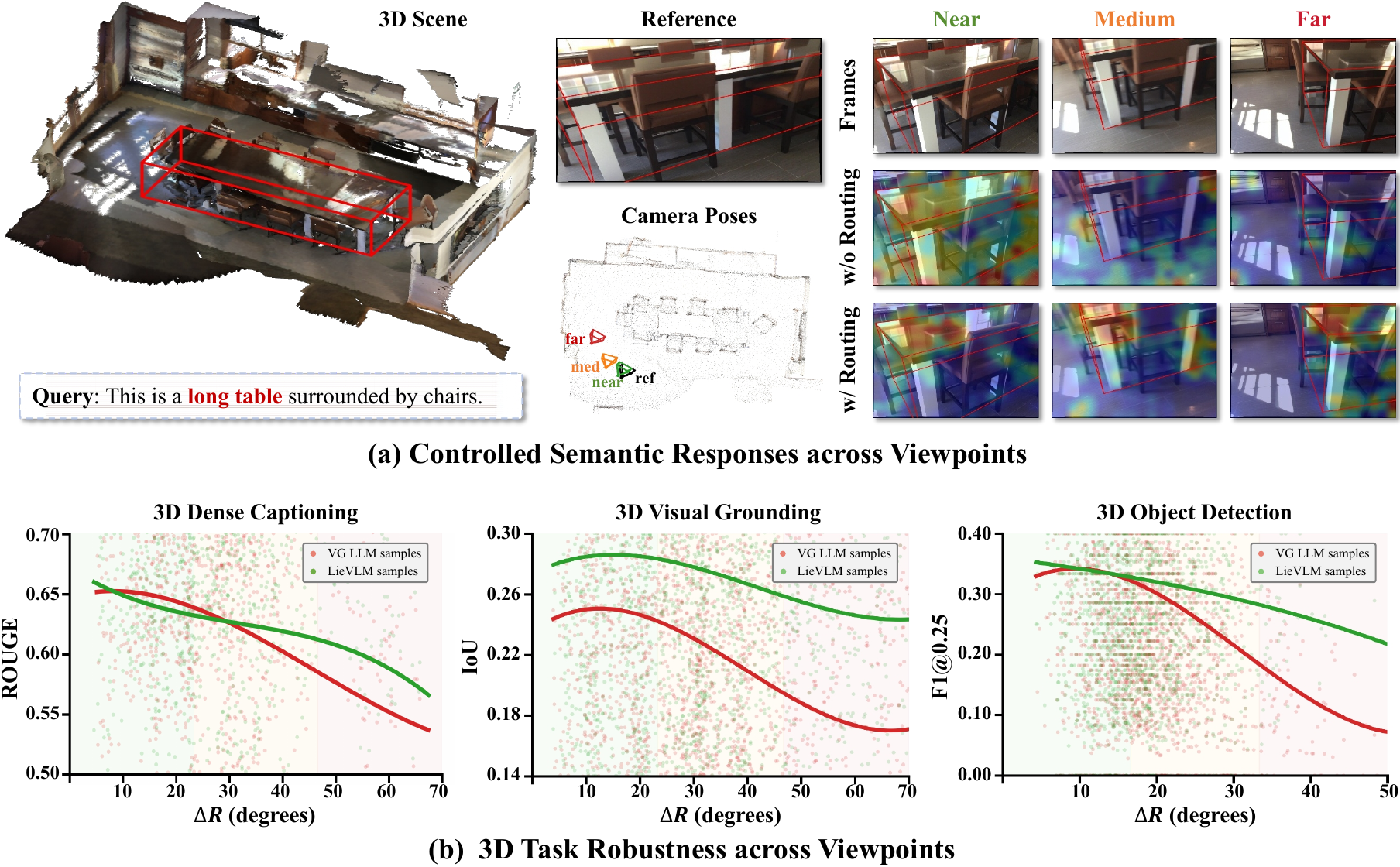}
    \vspace{-6mm}
    \caption{
    \textbf{3D scene viewpoint variation exposes semantic instability and downstream performance degradation.}
    (a) Without routing, query-relevant activations become diffuse and less target centered across views, while Lie routing preserves stable target responses. Warmer regions indicate stronger support for the target phrase.
    (b) Fitted task trends show that performance degrades
    as sample-level viewpoint variation $\Delta R$ increases, while LieVLM maintains stronger performance
    across higher levels of viewpoint variation. For grounding, raw IoU visualizes the continuous quantity underlying Acc@0.5.
    Additional examples are provided in Appendix~\ref{app:viewpoint_responses}.
    }
    \label{fig:teaser}
    \vspace{-6mm}
\end{figure}

We observe that a fixed scene entity and query can yield substantially different target-related responses across viewpoints. As camera pose, apparent scale, and visibility change, the response may weaken, become diffuse, or shift toward irrelevant regions, a phenomenon we term \emph{viewpoint-induced semantic drift}. We visualize it using query-relevant activations, obtained by projecting final visual tokens through the vocabulary head and averaging their positive support over the target-phrase tokens. Viewpoint change between frames $i$ and $j$ is measured by the $\mathrm{SO}(3)$ geodesic distance
$d_{\mathrm{rot}}(i,j)=\arccos\!\left((\operatorname{tr}(R_i^\top R_j)-1)/2\right)$~\citep{hartley2003multiple,huynh2009metrics}.
Taking the first frame containing the referred target as reference $r$, we define
$\Delta R=|V_s|^{-1}\sum_{t\in V_s}d_{\mathrm{rot}}(r,t)$
over the remaining views $V_s$, with both distances reported in degrees.
Figure~\ref{fig:teaser}(a) varies viewpoint along a fixed camera trajectory while holding the scene, target, query, checkpoint, and reference view constant, showing that Lie routing preserves more localized responses.
Figure~\ref{fig:teaser}(b) relates task performance to $\Delta R$, with LieVLM maintaining stronger performance as viewpoint variation increases.

To address this challenge, we introduce LieVLM, which adapts visual semantics through geometry-conditioned Lie routing.
Camera rotations are described by $\mathrm{SO}(3)$, motivating us to use the same structure for feature adaptation across viewpoints.
Lie routing applies learned rotations to three-dimensional groups of projected visual features, changing their directions while preserving their lengths.
Concretely, geometry tokens predict local $\mathfrak{so}(3)$ generators and channel-wise routing coefficients.
A Pad\'e--Cayley realization converts the generators into local rotations, and only the resulting identity-relative corrections are projected back into the MLLM representation.
The corrections are gated and fused with the original visual features and direct geometry features before language generation.
This design combines structured feature adaptation with retained semantic and spatial-context paths, without requiring relative camera poses as inputs to the routing module.

We evaluate LieVLM on 3D captioning, grounding, video object detection, and spatial reasoning~\citep{chen2021scan2cap,chen2020scanrefer,wang2024embodiedscan,yang2025thinking}. LieVLM consistently improves over matched additive geometry fusion, with gains that generalize across spatial benchmarks, training scales, and stronger multimodal backbones. Controlled analyses further connect these improvements to structured geometry conditioned transformations and robustness to viewpoint variation.

Our contributions are summarized as follows:
\begin{itemize}[itemsep=0.2pt, topsep=1pt, leftmargin=2em]
    \item We characterize viewpoint-induced semantic drift in 3D MLLMs through cross-view target-response diagnostics and its association with downstream task performance.
    \item We develop Lie Routing Vision Language Models (LieVLM), combining local $\mathfrak{so}(3)$ prediction, rotation-preserving residual transformations, and channel-wise gating to adapt pretrained visual semantics across viewpoints.
    \item We demonstrate consistent gains across 3D scene understanding and spatial reasoning, with controlled evaluations across training scales and backbones and targeted analyses of routing structure and viewpoint robustness.
\end{itemize}

\section{Related Work}
\label{sec:related}

\vspace{-1mm}
\paragraph{3D Vision Language Models.}
3D VLMs incorporate explicit spatial inputs such as point clouds and object representations~\citep{xu2024pointllm,chen2024ll3da,huang2024chat,zhu2024llava}, or derive geometric priors from image sequences~\citep{zheng2025learning,wang2025vggt,wang2024dust3r}. Reconstruction-aware models further couple video understanding with 3D reconstruction~\citep{chen2025vid,fan2025vlm,hu2025g}, establishing geometry as an important representation for 3D reasoning. Recent RGB-based systems additionally combine semantic and geometry encoders through camera-guided fusion~\citep{zhao2025spacemind}. Our work builds on these interfaces and studies geometry-conditioned routing of pretrained visual semantics across viewpoints.

\vspace{-3mm}
\paragraph{3D Scene Understanding and Spatial Reasoning.}
Scene understanding benchmarks evaluate localization, grounding, and captioning~\citep{chen2020scanrefer,chen2021scan2cap,wang2024embodiedscan}. Spatial reasoning benchmarks assess depth, distance, viewpoint, and multiview relations~\citep{yang2025thinking,tong2024cambrian,fu2024blink,zhang2026generalization}, while instruction datasets develop these capabilities from embodied observations~\citep{chen2024spatialvlm,zhang2025flatland,ray2024sat}. Multi-view grounding methods further aggregate view-dependent evidence through inter-view attention and learned view prototypes~\citep{guo2023viewrefer}. These efforts motivate our focus on semantic stability and task robustness under viewpoint variation.

\vspace{-3mm}
\paragraph{Geometry in Neural Representations.}
Prior work spans conditional feature modulation through FiLM~\citep{perez2018film}, explicit geometry-induced transformations in multi-view attention via GTA~\citep{miyato2024gta}, and equivariant representations under $\mathrm{SO}(3)$ or Euclidean symmetries~\citep{deng2021vector,hutchinson2021lietransformer,fuchs2020se,satorras2021en}. Orthogonal and Cayley parameterizations further provide structure-preserving learned transformations~\citep{helfrich2018orthogonal}. We instead study geometry-conditioned Lie routing in pretrained 3D MLLMs, where learned geometry tokens predict token-wise local $\mathrm{SO}(3)$ actions and channel-wise gates over latent subspaces for viewpoint-adaptive routing without explicit relative pose inputs.

\section{Method}
\label{sec:method_prelim}

\begin{figure*}[!t]
\centering
\includegraphics[width=1.0\textwidth]{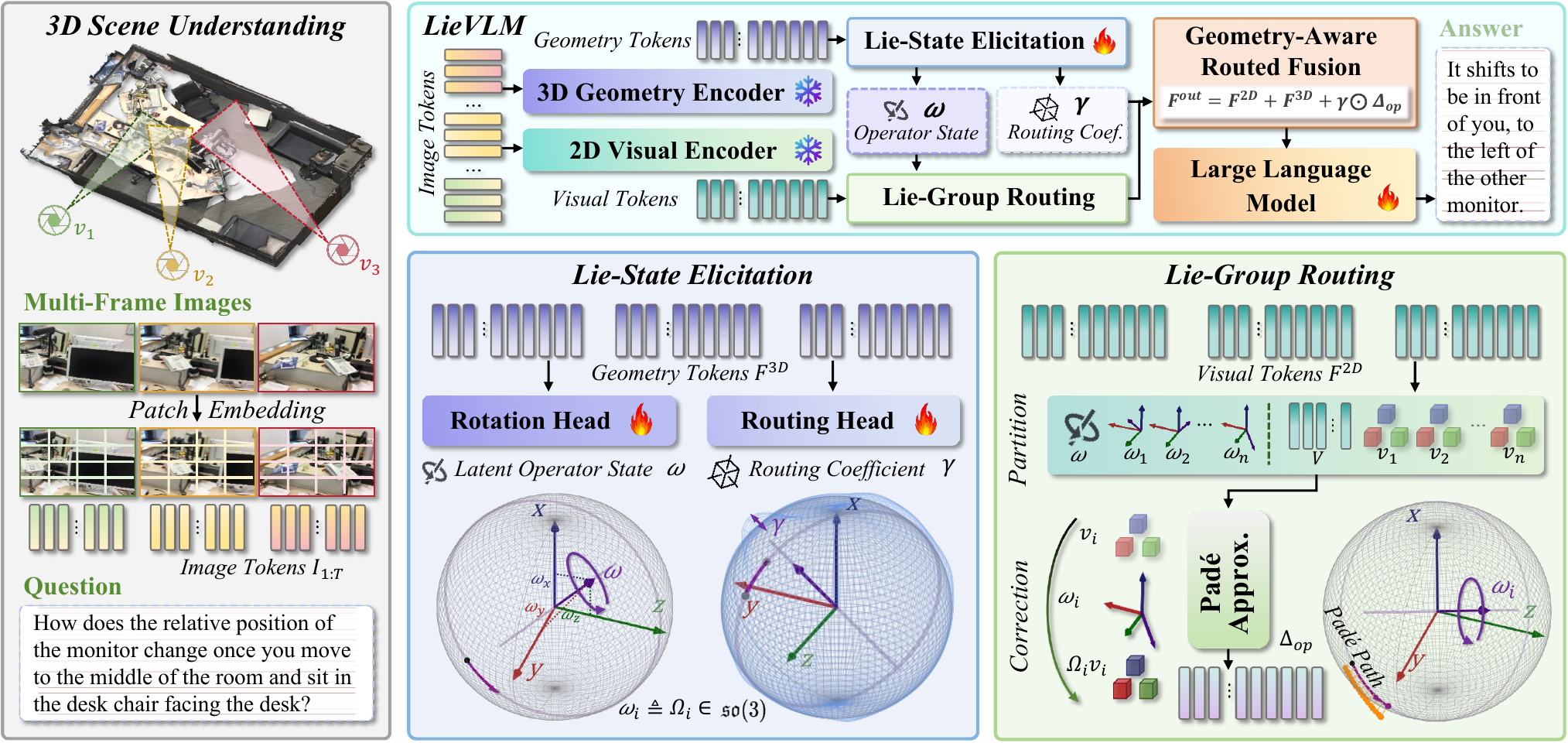}
\vspace{-5mm}
\caption{
\textbf{Framework of the proposed LieVLM.}
Lie-State Elicitation extracts latent operator states and routing coefficients from geometry tokens. Lie Group Routing transforms projected visual tokens, and routed fusion combines semantic, geometric, and operator paths before language generation.
}
\label{fig:model}
\vspace{-3mm}
\end{figure*}

\subsection{Preliminaries}
\label{sec:preliminary}

\vspace{-1mm}
\noindent \textbf{3D Geometry in MLLMs.}
Consider a video or multi-view observation $\{I_t\}_{t=1}^{T}$ of a 3D scene. A standard MLLM maps each frame into semantic visual tokens $F^{2d}_t\in\mathbb{R}^{N\times H}$, while the geometry encoder provides aligned geometry tokens $F^{3d}_t\in\mathbb{R}^{N\times H}$, where $H$ denotes the visual token width at the MLLM interface.
Existing 3D-enhanced VLMs commonly combine the two streams through additive or attention-based feature injection,
\begin{equation}
    \widetilde{F}_t = F^{2d}_t + B(F^{3d}_t),
    \label{eq:additive_fusion}
\end{equation}
where $B(\cdot)$ is a learned geometry projection. The 3D perception fundamentally concerns how scene observations transform across viewpoints. 
Beyond supplying spatial context, our goal is to adapt visual semantics to viewpoint variation through structured latent transformations.

\noindent \textbf{Lie Groups.}
We briefly review Lie groups, Lie algebras, and the Special Orthogonal group $\mathrm{SO}(3)$. For more in-depth details, please refer to~\citep{hall2013lie,hartley2003multiple}.

\begin{definition}[{\bf Lie Group}]
\label{def:lie_group}
A Lie group is a smooth manifold $G$ equipped with a group operation $\odot$ such that the multiplication map $m:(x,y)\mapsto x\odot y$ and the inverse map $i:x\mapsto x^{-1}$ are smooth.
\end{definition}

\begin{definition}[{\bf Lie Algebra}]
\label{def:lie_algebra}
Let $G$ be a Lie group with identity $e$. Its Lie algebra is the tangent space at $e$, \ie, $\mathfrak{g}=T_eG$, which locally linearizes $G$ through the exponential map $\exp:\mathfrak{g}\rightarrow G$.
\end{definition}

The set of 3D rotations, known as the Special Orthogonal group $\mathrm{SO}(3)$, forms a Lie group 
\begin{equation}
    \mathrm{SO}(3)=
    \{R\in\mathbb{R}^{3\times3}\mid R^\top R=I,\det(R)=1\},
    \quad
    \mathfrak{so}(3)=
    \{\Omega\in\mathbb{R}^{3\times3}\mid \Omega^\top=-\Omega\},
\end{equation}
where $\mathfrak{so}(3)$ is the Lie algebra of $\mathrm{SO}(3)$, namely the space of $3\times3$ skew-symmetric matrices. The standard hat operator $(\cdot)^\wedge:\mathbb{R}^3\rightarrow\mathfrak{so}(3)$ identifies an axis-angle vector $\omega$ with its skew-symmetric matrix $\Omega = \omega^\wedge$, whose exponential $\exp(\Omega)$ gives the corresponding rotation in $\mathrm{SO}(3)$.

\noindent \textbf{Lie Routing.}
Lie groups provide a natural framework for modeling geometric transformations. Since the geometry tokens $F^{3d}_{t}$ encode scene structure, viewpoint variation, and cross-view correspondence, they should not only serve as additional geometric context, but also indicate how semantic states should be adjusted under viewpoint changes. Conventional additive fusion injects a geometry-dependent feature, such as $B(F^{3d}_{t})$, into the visual representation. Although this enriches tokens with 3D information, it does not explicitly define an operation on the semantic state. 
Lie routing utilizes geometry features to predict local rotations of projected visual representations.
The routed branch retains only the change induced by these transformations relative to the identity,
\begin{equation}
F^{2d}_t
\mapsto
F^{2d}_t
+
\Delta_{\mathrm{op}}
\left(
F^{2d}_t
\,;\,
F^{3d}_t
\right).
\label{eq:residual_operator}
\end{equation}
Here, $F^{2d}_t$ denotes the semantic state and $F^{3d}_t$ provides the geometric conditioning. The term $\Delta_{\mathrm{op}}$ denotes only the semantic change induced by the operator, rather than the complete transformed state. By construction, this correction vanishes when the induced operator is the identity.
We implement this correction using local $\mathfrak{so}(3)$ generators and a rational $\mathrm{SO}(3)$ parameterization, preserving the pretrained semantic path while geometry governs viewpoint dependent semantic transformations.

\subsection{Lie-State Elicitation}
\label{sec:lie_state_elicitation}


For a video or multi-view observation $\{I_t\}_{t=1}^{T}$ of a 3D scene, VGGT~\citep{wang2025vggt} extracts 3D geometry features.
These features are aligned to the visual token grid and projected to the MLLM interface width $H$, yielding geometry tokens $F^{3d}\in\mathbb{R}^{N\times H}$.
Alignment details are provided in Appendix~\ref{app:imp}.
Layer normalization and two prediction heads then produce the latent rotation parameters and channel-wise routing coefficients,
\begin{equation}
Z=\operatorname{LN}(F^{3d}),\quad
\omega=\tanh(W_{\omega}Z)\in\mathbb{R}^{N\times D_o},\quad
\gamma=\sigma(W_{\gamma}Z+b_{\gamma})\in(0,1)^{N\times H}.
\label{eq:lie_state_prediction}
\end{equation}
Here, $W_{\omega}:\mathbb{R}^{H}\rightarrow\mathbb{R}^{D_o}$ predicts latent operator states, while $W_{\gamma}:\mathbb{R}^{H}\rightarrow\mathbb{R}^{H}$ predicts token and channel wise routing coefficients. The learnable routing bias $b_{\gamma}$ is initialized to $-4$, biasing $\gamma$ toward a low activation regime, since $\sigma(-4)\approx0.018$ when $W_{\gamma}Z=0$. We set $D_o=3\lfloor H/3\rfloor$ and define $M=D_o/3$ triplet groups, with $\gamma$ gating the routed transformation at each token and output channel.

To construct structured generators, we reshape the predicted state as
$\omega\in\mathbb{R}^{N\times M\times3}$,
with
$\omega_{n,m}=(\omega^x_{n,m},\omega^y_{n,m},\omega^z_{n,m})$,
where $n\in\{1,\ldots,N\}$ indexes visual tokens and $m\in\{1,\ldots,M\}$ indexes triplet subspaces. The partition is applied after projection to the operator space, rather than directly to the backbone hidden dimension. Each triplet defines a learned local latent subspace.
The triplet partition is defined over learned projected channels and assumes no predefined semantic ordering of channels.
Each $\omega_{n,m}$ parameterizes an element of $\mathfrak{so}(3)$ acting on the corresponding semantic triplet, with camera-dependent geometry provided by the aligned geometry tokens, and its exponential defines an $\mathrm{SO}(3)$ structured transformation in the latent subspace.

The latent operator state $\omega_{n,m}$ is mapped to a triplet-wise $\mathfrak{so}(3)$ generator through the hat map:
\begin{equation}
    \Omega_{n,m} =
    \begin{bmatrix}
        0 & -\omega^{z}_{n,m} & \omega^{y}_{n,m} \\
        \omega^{z}_{n,m} & 0 & -\omega^{x}_{n,m} \\
        -\omega^{y}_{n,m} & \omega^{x}_{n,m} & 0
    \end{bmatrix}
    \in \mathfrak{so}(3).
    \label{eq:hat_map}
\end{equation}
The geometry encoder produces triplet-wise $\mathfrak{so}(3)$ generators over the visual tokens.
The output is defined as the Lie state $S_{\mathrm{Lie}}=(\Omega,\gamma), \;
\Omega=(\Omega_{n,m})\in\mathfrak{so}(3)^{N\times M}$, where $\Omega_{n,m}$ specifies the local operator for the $m$-th latent group of token $n$, and $\gamma$ controls the routed contribution at each token.

\subsection{Lie-Group Routing}
\label{sec:lie_space_routing}

Given the Lie state $\mathcal{S}_{\mathrm{Lie}}$ elicited from geometry tokens, we use it to route the 2D semantic representation through geometry conditioned operators with $SO(3)$ structure. 
We first project the visual tokens into the operator space $V=W_v(\operatorname{LN}(F^{2d}))\in\mathbb{R}^{N\times D_o},$
where $W_v:\mathbb{R}^{H}\rightarrow\mathbb{R}^{D_o}$.
We then partition $V$ into $M=D_o/3$ semantic triplets
$\{v_{n,m}\in\mathbb{R}^{3}\}_{n,m}$, and pair each $v_{n,m}$ with its corresponding $\mathfrak{so}(3)$ generator $\Omega_{n,m}$.
The exponential map defines a standard $\mathrm{SO}(3)$ action for each generator and admits the Rodrigues closed form~\citep{murray2017mathematical}.
We adopt a diagonal Pad\'e realization~\citep{higham2008functions} that preserves the local rotation structure and yields a closed-form, rationally normalized correction.
Its structural properties and local sensitivity are analyzed in Appendix~\ref{app:theoretical_analysis}.
Its action is expressed through the first- and second-order generator terms:
\begin{equation}
    \label{eq:first_order_action}
    a_{n,m}=\Omega_{n,m}v_{n,m}=\omega_{n,m}\times v_{n,m},
\end{equation}
\begin{equation}
    b_{n,m}
    =
    {\Omega_{n,m}}^{2}v_{n,m}
    =
    \omega_{n,m}\times(\omega_{n,m}\times v_{n,m})
    =
    \omega_{n,m}\big(\omega_{n,m}^{\top}v_{n,m}\big)
    -
    \|\omega_{n,m}\|_2^2 v_{n,m}.
    \label{eq:second_order_action}
\end{equation}



Here, $a_{n,m}$ and $b_{n,m}$ are the first- and second- order generator actions on the semantic triplet. Let $R_{1,1}(\Omega_{n,m})$ denote the Pad\'e 1/1 approximation to $\exp(\Omega_{n,m})$, which is equivalent to the Cayley action reviewed in Appendix~\ref{app:so3_background}. LieVLM retains only the residual action relative to the identity,
\begin{equation}
c^{\mathrm{pade}}_{n,m}
=
\left(
R_{1,1}(\Omega_{n,m})-I
\right)
v_{n,m}
=
\frac{
a_{n,m}
+
\frac{1}{2}b_{n,m}
}{
1
+
\frac{1}{4}
\lVert\omega_{n,m}\rVert_2^2
}.
\label{eq:lie_routing}
\end{equation}
The corresponding transformed triplet is $v_{n,m}+c^{\mathrm{pade}}_{n,m}$, while only the correction $c^{\mathrm{pade}}_{n,m}$ enters the routed branch. 
We reshape all triplet-wise corrections into
$\delta^{\mathrm{pade}}\in\mathbb{R}^{N\times D_o}$.
A bias-free output projection maps the routed representation back to the MLLM token space,
\begin{equation}
\Delta_{\mathrm{op}}(F^{2d};F^{3d})
=
W_o(\delta^{\mathrm{pade}})
\in\mathbb{R}^{N\times H},
\label{eq:operator_correction}
\end{equation}
where $W_o:\mathbb{R}^{D_o}\rightarrow\mathbb{R}^{H}$.
The visual tokens $F^{2d}$ provides the content to be routed, while $F^{3d}$ conditions the $\mathfrak{so}(3)$ generators and routing strengths. This produces a geometry-conditioned operator correction that is later fused with the original semantic path and geometry anchor.

\subsection{Geometry-Aware Routed Fusion}
\label{sec:geometry_aware_routed_fusion}
The Lie Group Routing module produces the operator correction $\Delta_{\mathrm{op}}(F^{2d};F^{3d})$, which captures geometry-conditioned semantic transformation under the induced $\mathfrak{so}(3)$ generators. However, geometry tokens also contain 3D contextual information beyond rotational variation, such as scene layout, structural priors, and cross-view correspondence. We therefore fuse the original 2D semantic path with two complementary geometry paths: a direct 3D context path and a Lie-routed transformation path. The final geometry-aware visual representation is computed as
\begin{equation}
    F^{out}
    =
    F^{2d}
    +
    F^{3d}
    +
    \gamma \odot \Delta_{\mathrm{op}}(F^{2d};F^{3d}),
    \label{eq:routed_fusion}
\end{equation}
where $\odot$ denotes element-wise multiplication and
$\gamma\in(0,1)^{N\times H}$ is the token- and channel-wise routing
coefficient predicted from the geometry tokens in Eq.~\ref{eq:lie_state_prediction}.
The three terms in Eq.~\ref{eq:routed_fusion} play complementary roles. The first term $F^{2d}$ preserves the pretrained 2D visual-semantic representation. The second term $F^{3d}$ provides direct 3D contextual information, and
$\gamma\odot\Delta_{\mathrm{op}}(F^{2d};F^{3d})$
introduces a geometry-conditioned correction to the semantic representation. The direct geometry path keeps 3D evidence available to the language model, while the Lie-routed path uses the geometry-conditioned operators from Sec.~\ref{sec:lie_space_routing} to model transformation-aware semantic variation.
The routing coefficient $\gamma$ adaptively controls the Lie routed contribution for each token and channel.

This fusion design preserves the pretrained semantics and direct geometry paths throughout optimization, while the bounded routing coefficient $\gamma$ modulates only the Lie-routed transformation. The routing head is learned jointly with the task objective, allowing the transformation strength to adapt across visual tokens and output channels.

Finally, $F^{out}$ is used as the visual representation passed to the MLLM backbone for language generation. For a video or multi-view input, the same fusion rule is applied to each frame's aligned token grid, and the resulting fused visual tokens are concatenated with text embeddings following the standard MLLM interface. The model is trained end-to-end with the task objective without requiring explicit supervision on $\omega$, $\Omega$, or $\gamma$.

\vspace{-1.5mm}
\section{Experiments}

\vspace{-1mm}
\subsection{3D Scene Understanding} \label{sec:3d_exp}

\vspace{-2mm}
\paragraph{Datasets \& Evaluation Protocols.}
We evaluate LieVLM on 3D visual grounding, 3D dense captioning, and 3D video object detection. 
For grounding, we use the ScanRefer dataset ~\citep{chen2020scanrefer} under the protocol of SPAR~\citep{zhang2025flatland}, with Mask3D proposal refinement applied consistently to methods SPAR, VG LLM, and LieVLM.
For dense captioning, we use the Scan2Cap dataset~\citep{chen2021scan2cap} and generate object descriptions conditioned on Mask3D proposals from LEO~\citep{huang2023embodied}, following prior work~\citep{huang2023embodied,zheng2025video,zhu2024llava}. For 3D video object detection, we adopt the VG LLM protocol on the EmbodiedScan dataset~\citep{wang2024embodiedscan}.
We report Acc@0.25/0.5 for ScanRefer, CIDEr, BLEU-4, METEOR, and ROUGE under the @0.5 criterion for Scan2Cap, and precision, recall, and F1 at IoU 0.25 over 20 common classes for detection.
All models are trained for one epoch with Adam, a batch size of 64, a warmup ratio of 0.03, and a peak learning rate of $1\times10^{-5}$ followed by linear decay. 
The 2D visual encoder, 3D geometry encoder, and multimodal connector are frozen, while the learnable routing bias is initialized to $b_\gamma=-4$ and fixed as the default across all main experiments.

\vspace{-3.5mm}
\paragraph{Baselines.}
The baselines are grouped according to how 3D geometry is incorporated. For ScanRefer and Scan2Cap, we compare with explicit 3D-input methods, including ScanRefer~\citep{chen2020scanrefer}, Scan2Cap~\citep{chen2021scan2cap}, 3D-LLM~\citep{hong20233d}, LL3DA~\citep{chen2024ll3da}, Chat-3D v2~\citep{huang2023chat}, Grounded 3D-LLM~\citep{chen2024grounded}, LEO~\citep{huang2023embodied}, Chat-Scene~\citep{huang2024chat}, LLaVA-3D~\citep{zhu2024llava}, and Video-3D LLM~\citep{zheng2025video}. We also include geometry-injected MLLMs, including SPAR~\citep{zhang2025flatland}, 3DRS~\citep{huang20253drs}, and VG LLM~\citep{zheng2025learning}. For 3D video object detection, we compare against Qwen2.5-VL~\citep{bai2025qwen25vltechnicalreport} and VG LLM~\citep{zheng2025learning}. 
Tables~\ref{tab:scan2cap}--\ref{tab:3d_det_exp} use published VG LLM results, while controlled analyses use our reproduced 8B baselines.

\vspace{-3.5mm}
\paragraph{Results on 3D Dense Captioning.}
Table~\ref{tab:scan2cap} reports the Scan2Cap results. LieVLM achieves the best CIDEr score of 82.7, surpassing Video-3D LLM by 2.7 points and LLaVA-3D by 3.5 points. Compared with geometry-injected RGB/video baselines, LieVLM improves over VG LLM-8B from 80.0 to 82.7 in CIDEr while remaining competitive across the other captioning metrics. 
These results support the effectiveness of Lie routing for object-level captioning.

\begin{figure}[!t]
\centering
\includegraphics[width=\textwidth]{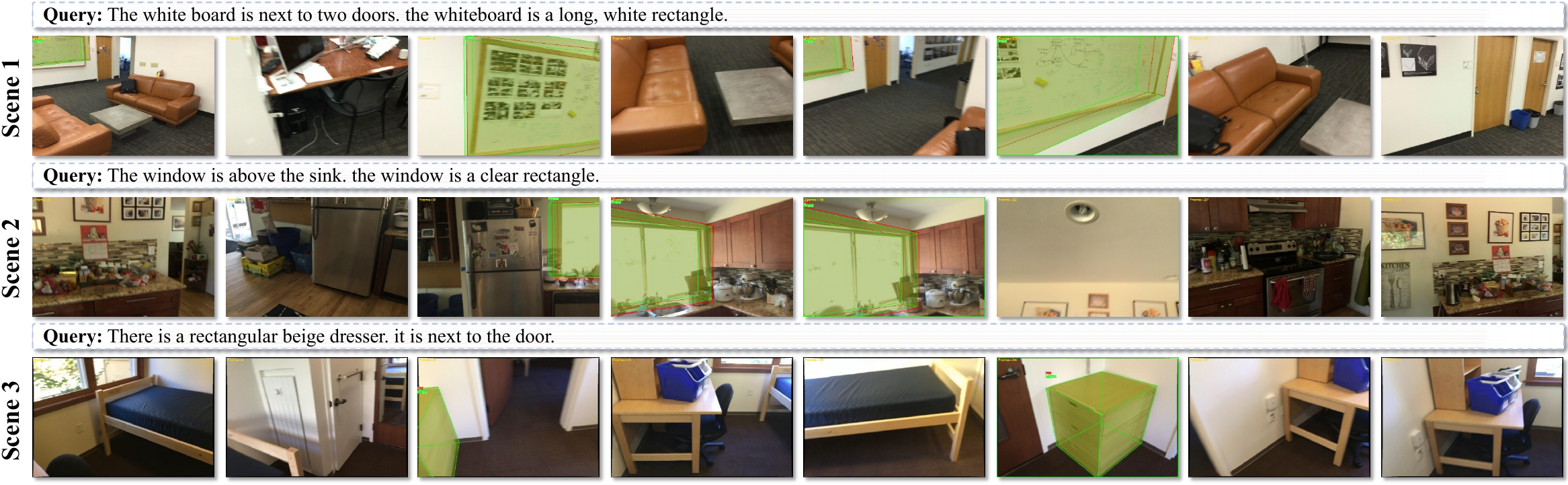}
\vspace{-6.5mm}
\caption{Qualitative results of LieVLM for 3D visual grounding.}
\label{fig:vg_visualization}
\vspace{-3mm}
\end{figure}

\vspace{-3.5mm}
\paragraph{Results on 3D Visual Grounding.}
Table~\ref{tab:scanrefer} shows the ScanRefer results. LieVLM reaches 64.5 Acc@0.25 and 57.2 Acc@0.5. Under the same proposal refinement protocol, it improves over the published VG LLM-8B results by 6.9 and 6.3 points, respectively, demonstrating stronger grounding performance with geometry conditioned operator routing.
Results in Fig.~\ref{fig:vg_visualization} show that the predictions closely overlap with the ground-truth boxes, demonstrating accurate language-guided 3D localization.

\begin{table}[!t]
\begin{minipage}[t]{0.58\textwidth}
\caption{The quantitative results of 3D dense captioning on Scan2Cap~\citep{chen2021scan2cap}.
Underline denotes overall best result, and bold denotes best results among RGB-based methods.
}
\resizebox{\linewidth}{!}{
\begin{tabular}{lcccc}
\toprule
\makecell[c]{Method} & C@0.5$\uparrow$ & B-4@0.5$\uparrow$ & M@0.5$\uparrow$ & R@0.5$\uparrow$ \\
\hline
\rowcolor{gray!15}
\multicolumn{5}{l}{\textit{Explicit 3D Input}} \\
Scan2Cap~\citep{chen2021scan2cap} & 39.1 & 23.3   & 22.0 & 44.8 \\
LL3DA~\citep{chen2024ll3da} & 65.2  & 36.8 & 26.0& 55.0   \\
Chat-3D-v2 \citep{huang2023chat} & 63.9 & 31.8 & - & - \\
Grounded 3D-LLM \citep{chen2024grounded} & 70.2 & 35.0 & - & - \\
LEO~\citep{huang2023embodied} & 72.4 &38.2 &27.9 &58.1 \\ 
Chat-Scene~\citep{huang2024chat} & 77.1 &36.5 &- &- \\
LLaVA-3D~\citep{zhu2024llava} & 79.2 & 41.1 & \underline{30.2} & \underline{63.4} \\
Video-3D LLM~\citep{zheng2025video} & 80.0 & 40.2 & 28.5 & 61.7 \\
\midrule
\rowcolor{gray!15}
\multicolumn{5}{l}{\textit{RGB-Based Models}} \\
3DRS \citep{huang20253drs} & 76.4 & 39.6 & 27.3 & 57.1 \\
VG LLM-4B~\citep{zheng2025learning}
& 78.6 & 40.9 & 28.6 & 62.4 \\

VG LLM-8B~\citep{zheng2025learning}
& 80.0 & 41.5 & 28.9 & 62.6 \\
    \textbf{LieVLM-8B (Ours)} & \underline{\textbf{82.7}} & \underline{\textbf{41.7}} & \textbf{28.9} & \textbf{62.9} \\
\bottomrule
\end{tabular}
}
\label{tab:scan2cap}
\end{minipage}
\hfill
\begin{minipage}[t]{0.3923\textwidth}
\centering
\caption{The results of 3D Visual grounding on ScanRefer.~\citep{chen2020scanrefer}.
Underline denotes overall best, and bold denotes best among RGB-based methods.
} 
\resizebox{\linewidth}{!}{
\begin{tabular}{lcc}
  \toprule
  \makecell[c]{Model} & Acc@0.25 & Acc@0.5 \\
  \midrule
  \rowcolor{gray!15}
  \multicolumn{3}{l}{\textit{Explicit 3D Input}} \\
  ScanRefer ~\citep{chen2020scanrefer} & 37.3 & 24.3 \\
  Chat-3D v2~\citep{huang2023chat} & 35.9 & 30.4 \\
  Grounded 3D-LLM~\citep{chen2024grounded} & 47.9 & 44.1 \\ 
  ChatScene~\citep{huang2024chat} & 55.5 & 50.2 \\
  LLaVA-3D \citep{zhu2024llava} & 54.1 & 42.4 \\
  Video-3D LLM \citep{zheng2025video} & 58.1 & 51.7 \\ 
  \midrule
  \rowcolor{gray!15}
  \multicolumn{3}{l}{\textit{RGB-Based Models}} \\
  SPAR~\citep{zhang2025flatland} & 48.8 & 43.1 \\
  3DRS \citep{huang20253drs} & 55.2 & 51.1 \\
  VG LLM-4B~\citep{zheng2025learning}
  & 53.5 & 47.5 \\
    
  VG LLM-8B~\citep{zheng2025learning}
  & 57.6 & 50.9 \\
  \textbf{LieVLM-8B (Ours)} & \underline{\textbf{64.5}} & \underline{\textbf{57.2}}  \\
  \bottomrule
\end{tabular}
}
\label{tab:scanrefer}
\end{minipage}
\vspace{1mm}
\caption{The quantitative results on 3D video object detection on EmbodiedScan~\citep{wang2024embodiedscan}.} 
\vspace{-2.7mm}
\resizebox{\linewidth}{!}{
    \begin{tabular}{l|cccccccc|ccc}
    \toprule
    \makecell[c]{\multirow{2}*{Method}} & \multirow{2}*{chair} & \multirow{2}*{cabinet} & \multirow{2}*{table} & \multirow{2}*{bin} & \multirow{2}*{couch} & \multirow{2}*{bed} & \multirow{2}*{bathtub} & \multirow{2}*{toilet}  & \multicolumn{3}{c}{20 Common Classes} \\
    \cmidrule{10-12}
    &   & ~ & ~ & ~ & ~ & ~ & ~ & ~ & Prec.@0.25 & Recall@0.25 & F1@0.25\\
    \midrule
    \rowcolor{gray!15}
    \multicolumn{12}{l}{\textit{4-Frame Setting}} \\ 
    Qwen2.5-VL \citep{bai2025qwen25vltechnicalreport} & 41.2 & 11.6 & 36.5 & 30.2 & 41.1 & 68.2 & 36.6 & 68.7 & 34.6 & 31.0 & 32.5 \\
    VG LLM-8B \citep{zheng2025learning} & \textbf{54.0} & 17.1 & 46.5 & 39.8 & 47.0 & 74.1 & 42.1 & 82.5  & 43.4 & 39.6 & 41.2 \\
    \textbf{LieVLM-8B (Ours)} & 53.5 & \textbf{21.3} & \textbf{56.8} & \textbf{43.0} & \textbf{55.1} & \textbf{82.9} & \textbf{75.9} & \textbf{86.2} & \textbf{48.6} & \textbf{44.1} & \textbf{45.8} \\
    \hline
    \rowcolor{gray!15}
    \multicolumn{12}{l}{\textit{6-Frame Setting}} \\ 
    Qwen2.5-VL \citep{bai2025qwen25vltechnicalreport} & 36.1 & 10.6 & 32.7 & 25.0 & 40.7 & 64.6 & 38.4 & 68.6 & 31.8 & 28.0 & 29.6 \\
    VG LLM-8B \citep{zheng2025learning} & 48.7 & 17.9 & 44.8 & \textbf{38.5} & 46.4 & 75.8 & 40.4 & 83.2  & 43.5 & 38.7 & 40.0 \\
    \textbf{LieVLM-8B (Ours)} & \textbf{50.0} & \textbf{19.6} & \textbf{54.2} & 35.5 & \textbf{51.7} & \textbf{83.7} & \textbf{79.9} & \textbf{86.5} & \textbf{47.9} & \textbf{41.9} & \textbf{44.1} \\
    \bottomrule
    \end{tabular}
    }
\label{tab:3d_det_exp}
\vspace{-2mm}
\end{table}

\vspace{-3.5mm}
\paragraph{Results on 3D Video Object Detection.}
Table~\ref{tab:3d_det_exp} evaluates 3D video object detection under the 4-frame and 6-frame settings. LieVLM improves F1 from 41.2 to 45.8 in the 4-frame setting and from 40.0 to 44.1 in the 6-frame setting. 
Matched three-seed results in Table~\ref{tab:mechanism_ablation} further confirm consistent improvements, with detailed statistics in Appendix~\ref{app:embodiedscan}.
The category-level effects vary across objects, with particularly strong gains on table, bed and bathtub.

\begin{figure}[!t]
\centering
\includegraphics[width=\textwidth]{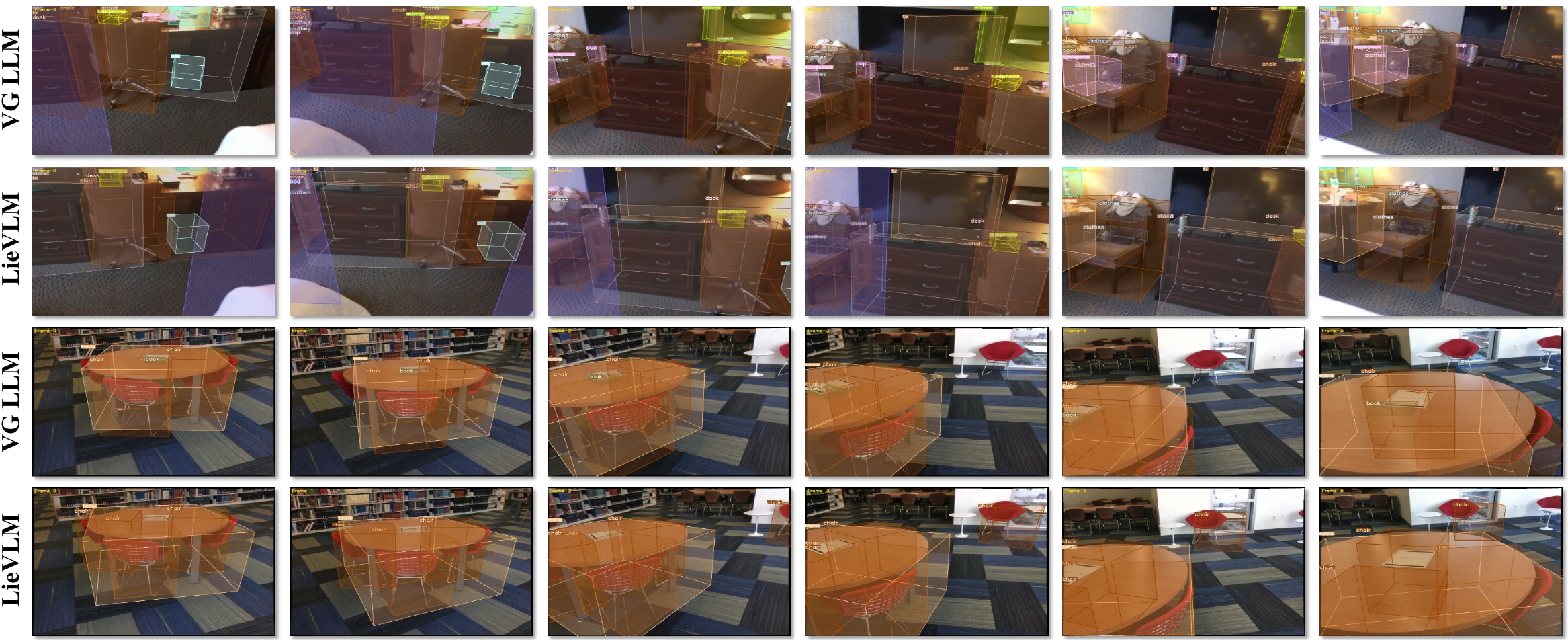}
\vspace{-6.5mm}
\caption{Qualitative results for 3D video object detection.}
\label{fig:visualization}
\vspace{-5mm}
\end{figure}

\subsection{General Spatial Reasoning}
\label{sec:general_spatial_reasoning}

\vspace{-0.8mm}
\paragraph{Datasets \& Evaluation Protocols.}
We train on a 3.05M example generalist mixture constructed from the public data sources released with CA-MLLM~\citep{zhang2026generalization}, covering video instruction following, spatial question answering, 3D grounding and captioning, 3D detection, and object centric spatial reasoning. We evaluate Qwen2.5-VL-7B variants on SPAR-Bench Full~\citep{zhang2025flatland}, CV-Bench-3D~\citep{tong2024cambrian}, and BLINK-Spatial~\citep{fu2024blink}, and Qwen3-VL-8B variants on VSI-Bench~\citep{yang2025thinking}, without benchmark-specific fine tuning. SPAR-Bench reports difficulty and camera sensitive categories, CV-Bench-3D and BLINK-Spatial report accuracy by spatial subtask, and VSI-Bench combines numerical Mean Relative Accuracy with multiple choice accuracy. The training composition is provided in Appendix~\ref{app:large_scale_spatial}.

\vspace{-3mm}
\paragraph{Baselines.}
We compare with representative generalist MLLMs and spatially specialized models, and include SenseNova-SI~\citep{cai2026scaling} as an external Qwen3 scale reference. Within each controlled comparison, VG LLM and LieVLM share the same language and geometry backbones, training data, optimization, and evaluation, and differ only in additive geometry fusion or Lie routing.

\begin{table*}[t]
\centering
\scriptsize
\setlength{\tabcolsep}{3.1pt}
\caption{
Comparison on SPAR-Bench Full under the 3.05M setting.
Prior model results follow published evaluations, and controlled rows use identical training and evaluation.
}
\vspace{-2.5mm}
\label{tab:spar_full}
\resizebox{\textwidth}{!}{
\begin{tabular}{lcccccccccc}
\toprule
\textbf{Method}
& \makecell{\textbf{Data}\\\textbf{Size}}
& \textbf{Avg.}
& \textbf{Low}
& \textbf{Medium}
& \textbf{High}
& \textbf{CamMotion}
& \textbf{ViewChgI}
& \textbf{DistI-OO-MV}
& \textbf{ObjRel-OC-MV}
& \textbf{SpImag-OO-MV} \\
\midrule

\rowcolor{gray!15}
\multicolumn{11}{l}{\textit{Prior MLLMs and Spatial Models}} \\

InternVL2.5-8B~\citep{chen2024expanding}
& N/A
& 36.28
& 29.46
& 31.88
& 43.80
& 28.00
& 6.32
& 56.85
& 51.75
& 39.50 \\

LLaVA-OV-7B~\citep{li2024llava}
& N/A
& 31.20
& 21.79
& 26.13
& 40.14
& 30.25
& 9.47
& 55.06
& 37.25
& 35.01 \\

Qwen2.5-VL-7B~\citep{bai2025qwen25vltechnicalreport}
& N/A
& 33.07
& 28.75
& 22.97
& 40.27
& 28.75
& 6.83
& 51.49
& 44.75
& 31.93 \\

SPAR-8B~\citep{zhang2025flatland}
& 2M
& 63.25
& 65.53
& 63.01
& 61.29
& 73.00
& 37.14
& 79.76
& 64.00
& 53.50 \\

VG LLM-4B~\citep{zheng2025learning}
& 297K
& 60.36
& 52.81
& 51.35
& 72.92
& 63.00
& 25.82
& 63.69
& 81.25
& 71.99 \\

CA-MLLM-4B~\citep{zhang2026generalization}
& 3.05M
& 68.35
& 59.94
& 60.42
& 81.74
& 73.00
& 31.01
& 68.75
& 87.50
& 81.23 \\

\midrule
\rowcolor{gray!15}
\multicolumn{11}{l}{\textit{Controlled 3.05M Setting}} \\

VG LLM-8B~\citep{zheng2025learning}
& 3.05M
& \underline{67.18}
& 59.08
& 59.31
& 80.07
& 70.18
& 32.11
& 70.42
& 84.93
& 77.65 \\

\textbf{LieVLM-8B (Ours)}
& 3.05M
& \textbf{68.57}\,{\textcolor{green!50!black}{\tiny +1.39}}
& 59.81\,{\textcolor{green!50!black}{\tiny +0.73}}
& 60.74\,{\textcolor{green!50!black}{\tiny +1.43}}
& 81.21\,{\textcolor{green!50!black}{\tiny +1.14}}
& 70.92\,{\textcolor{green!50!black}{\tiny +0.74}}
& 35.26\,{\textcolor{green!50!black}{\tiny +3.15}}
& 71.58\,{\textcolor{green!50!black}{\tiny +1.16}}
& 85.34\,{\textcolor{green!50!black}{\tiny +0.41}}
& 78.27\,{\textcolor{green!50!black}{\tiny +0.62}} \\

\bottomrule
\end{tabular}}
\vspace{-2.7mm}
\end{table*}

\begin{table*}[t]
\centering
\begin{minipage}[t]{0.48\textwidth}
\centering
\scriptsize
\setlength{\tabcolsep}{2.5pt}
\caption{
Comparison on CV-Bench-3D and BLINK-Spatial under the 3.05M setting.
}
\vspace{-2.5mm}
\label{tab:cv_blink}
\resizebox{0.95\linewidth}{!}{
\begin{tabular}{lc|ccc|cccc}
\toprule
\multirow{2}{*}{\textbf{Method}}
& \multirow{2}{*}{\makecell{\textbf{Data}\\\textbf{Size}}}
& \multicolumn{3}{c|}{\textbf{CV-Bench-3D}}
& \multicolumn{4}{c}{\textbf{BLINK-Spatial}} \\
\cmidrule(lr){3-5}
\cmidrule(l){6-9}
&
& \textbf{Avg.}
& \textbf{Depth}
& \textbf{Dist.}
& \textbf{Avg.}
& \textbf{Depth}
 & \textbf{Spatial}
& \textbf{Multi View} \\
\midrule

\rowcolor{gray!15}
\multicolumn{9}{l}{\textit{Prior MLLMs and Spatial Models}} \\

GPT-4o
& N/A
& 83.0 & - & -
& 67.6 & 74.2 & 69.2 & 59.4 \\

Gemini-2.5-Pro
& N/A
& 90.8 & 91.0 & 90.7
& 71.6 & \textbf{87.9} & \textbf{91.6} & 35.3 \\

Qwen2.5-VL-7B
& N/A
& 77.3 & 84.5 & 76.5
& 73.8 & 79.0 & \underline{86.7} & 55.6 \\

Qwen2.5-VL-72B
& N/A
& 81.0 & 88.7 & 81.0
& 68.6 & 79.8 & 85.3 & 40.6 \\

SPAR-8B
& 2M
& 89.1 & - & -
& - & - & - & - \\

VG LLM-4B
& 297K
& 91.3 & - & -
& 68.4 & 79.8 & 71.3 & 54.1 \\

CA-MLLM-4B
& 3.05M
& 90.7 & 89.8 & 91.5
& \underline{77.0} & 77.4 & 66.4 & \textbf{87.2} \\

\midrule
\rowcolor{gray!15}
\multicolumn{9}{l}{\textit{Controlled 3.05M Setting}} \\

VG LLM-8B
& 3.05M
& \underline{92.6}
& \underline{92.1}
& \underline{93.1}
& 75.9
& 83.5
& 77.4
& 66.6 \\

\textbf{LieVLM-8B (Ours)}
& 3.05M
& \textbf{94.3}
& \textbf{93.9}
& \textbf{94.6}
& \textbf{77.1}
& \underline{84.0}
& 78.1
& \underline{69.1} \\[-0.70ex]
& &
{\textcolor{green!50!black}{\tiny +1.7}}
& {\textcolor{green!50!black}{\tiny +1.7}}
& {\textcolor{green!50!black}{\tiny +1.6}}
& {\textcolor{green!50!black}{\tiny +1.2}}
& {\textcolor{green!50!black}{\tiny +0.5}}
& {\textcolor{green!50!black}{\tiny +0.7}}
& {\textcolor{green!50!black}{\tiny +2.5}}
\\[-0.45ex]
\bottomrule
\end{tabular}}
\end{minipage}
\hfill
\begin{minipage}[t]{0.5033\textwidth}
\centering
\scriptsize
\setlength{\tabcolsep}{2.6pt}
\caption{
VSI-Bench comparison across backbone and scales with multiple choice categories.
}
\vspace{-2.5mm}
\label{tab:vsi_scaling}
\resizebox{\linewidth}{!}{
\begin{tabular}{lc|cccc|c}
\toprule
\multirow{2}{*}{\textbf{Method}}
& \multirow{2}{*}{\makecell{\textbf{Data}\\\textbf{Size}}}
& \multicolumn{4}{c|}{\textbf{Multiple Choice}}
& \multirow{2}{*}{\makecell{\textbf{Overall}\\\textbf{Avg.}}} \\
\cmidrule(lr){3-6}
&
& \textbf{Rel. Dist.}
& \textbf{Rel. Dir.}
& \textbf{Route}
& \textbf{Appr.}
& \\
\midrule

\rowcolor{gray!15}
\multicolumn{7}{l}{\textit{Prior MLLMs and Spatial Models}} \\

VG LLM-4B
& 297K
& 44.6
& 45.6
& 33.5
& 36.4
& 47.3 \\

CA-MLLM-4B
& 3.05M
& 50.7
& 45.5
& 31.4
& 20.2
& 46.8 \\

Qwen3-VL-8B
& N/A
& -
& -
& -
& -
& 57.9 \\

SenseNova-SI\,\scalebox{0.65}{Qwen3-VL-8B}
& 8M
& -
& -
& -
& -
& \textbf{64.8} \\

\midrule
\rowcolor{gray!15}
\multicolumn{7}{l}{\textit{Controlled Lie Routing Comparisons}} \\

VG LLM-8B
& 297K
& 46.6
& 40.7
& 32.4
& 59.2
& 50.6 \\

\textbf{LieVLM-8B (Ours)}
& 297K
& 49.1\,{\textcolor{green!50!black}{\tiny +2.5}}
& 42.9\,{\textcolor{green!50!black}{\tiny +2.2}}
& 33.5\,{\textcolor{green!50!black}{\tiny +1.1}}
& 59.8\,{\textcolor{green!50!black}{\tiny +0.6}}
& 52.4\,{\textcolor{green!50!black}{\tiny +1.7}} \\

VG LLM-Qwen3
& 3.05M
& \underline{57.8}
& \underline{52.2}
& \underline{45.0}
& \underline{69.9}
& 60.6 \\

\textbf{LieVLM-Qwen3 (Ours)}
& 3.05M
& \textbf{60.6}\,{\textcolor{green!50!black}{\tiny +2.8}}
& \textbf{55.5}\,{\textcolor{green!50!black}{\tiny +3.3}}
& \textbf{48.8}\,{\textcolor{green!50!black}{\tiny +3.8}}
& \textbf{73.7}\,{\textcolor{green!50!black}{\tiny +3.8}}
& \underline{62.8}\,{\textcolor{green!50!black}{\tiny +2.2}} \\

\bottomrule
\end{tabular}}
\end{minipage}
\vspace{-4.5mm}
\end{table*}

\vspace{-3mm}
\paragraph{Generalization across Benchmarks.}
Tables~\ref{tab:spar_full} and~\ref{tab:cv_blink} show consistent matched gains across every reported metric on SPAR-Bench Full, CV-Bench-3D, and BLINK-Spatial. LieVLM reaches the highest SPAR-Bench overall score with its largest gain on viewpoint change inference, and achieves the strongest CV-Bench-3D results across both depth and distance. The improvement further transfers to BLINK-Spatial, 
providing evidence beyond the original ScanNet centered evaluation suite.

\vspace{-3mm}
\paragraph{Scalability across Backbones and Data.}
Table~\ref{tab:vsi_scaling} shows that the matched benefit of Lie routing persists from the original Qwen2.5 setting to the scaled Qwen3 setting. With Qwen3-VL-8B and the 3.05M mixture, LieVLM raises the VSI-Bench average from 60.6 to 62.8 and improves seven of eight categories, including all four multiple choice categories shown in the table. The 8M SenseNova-SI results provide additional scale context, while the controlled Qwen3 comparison shows that the benefit of Lie routing persists with a stronger backbone and 10.3 times more training data. Full VSI-Bench category results and multi-seed statistics are provided in Appendix~\ref{app:vsi_details}.

\vspace{-1.5mm}
\subsection{Understanding Lie Routing}
\label{sec:operator_analysis}

\begin{figure*}[!t]
    \centering
    \includegraphics[width=\textwidth]{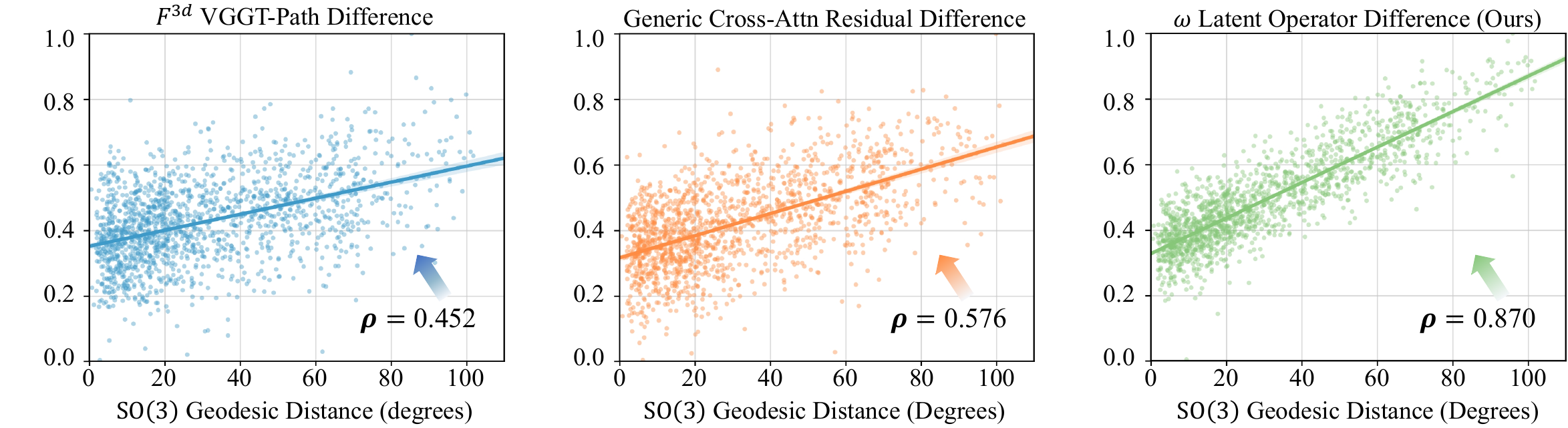}
    \vspace{-5.8mm}
    \caption{Routing viewpoint correspondence. Spearman correlation $\rho$ between physical camera rotation and internal representation change across same scene frame pairs.}
    \label{fig:operator_geometry}
    \vspace{-3.2mm}
\end{figure*}

\vspace{-1mm}
\paragraph{Routing Viewpoint Correspondence.}
A premise of Lie routing is that latent transformations should vary systematically with viewpoint while preserving target-related semantics. 
For same-scene frame pairs, with tokens matched by shared indices on the aligned visual grid, we relate the $\mathrm{SO}(3)$ camera rotation distance $d_{\mathrm{rot}}$ to the mean $\ell_2$ change in within-space normalized geometry features, cross-attention residuals, and latent operator states.
As shown in Figure~\ref{fig:operator_geometry}, 
operator state variation shows the strongest monotonic association with camera rotation among the representations considered, reaching Spearman $\rho=0.870$ compared with $0.452$ for geometry features and $0.576$ for cross-attention residuals. 
The stronger rank association indicates that the predicted transformation states vary systematically with viewpoint under this diagnostic.
This characterizes the adaptive behavior of Lie routing and complements the target-localization and task-level evidence in Figure~\ref{fig:teaser}.

\vspace{-3.2mm}
\paragraph{Operator Structure.}
To disentangle transformation structure, conditioning source, and response control, we compare variants under the same backbone, direct geometry path, training, and evaluation, with implementation details in Appendix~\ref{app:imp}. 
\emph{Additive Only} removes routing branch, while \emph{Visual-Conditioned Routing} predicts the operator state and gate from $F^{2d}$ instead of $F^{3d}$. \emph{Generic cross-attention} replaces the structured transformation with cross-attention from $F^{2d}$ to $F^{3d}$, and \emph{Learnable $3{\times}3$ Generator} applies an unconstrained geometry-conditioned matrix $G_{n,m}$ through $c^{\mathrm{gen}}_{n,m}=G_{n,m}v_{n,m}$, with spectral- and rational-controlled variants isolating response regulation. As shown in Table~\ref{tab:mechanism_ablation}, cross-attention yields modest gains and the unconstrained generator remains comparable to additive fusion, while spectral and rational controls provide progressive but limited improvements. Visual-conditioned routing likewise recovers only part of the gain. The geometry-conditioned $\mathfrak{so}(3)$--Pad\'e formulation remains consistently strongest, supporting the joint role of geometric conditioning and structured operator realization.

\vspace{-3.5mm}
\paragraph{Operator Realization and Routing.}
We further examine the numerical realization of the structured operator and its routing bias initialization. The Rodrigues baseline uses the closed-form action with small-angle series for coefficient evaluation near the identity, providing an exact exponential-action reference.
As shown in Tables~\ref{table:ablation1} and~\ref{table:ablation2}, the Pad\'e variants achieve the strongest overall results among the evaluated realizations. Pad\'e 1/1 gives the highest CIDEr and detection F1, while $b_{\gamma}=-4$ yields the best CIDEr and detection precision and $b_{\gamma}=-3$ gives higher recall. 
Appendix~\ref{app:theoretical_analysis} characterizes its rotation structure, response normalization, and local sensitivity.


\begin{table*}[!t]
\centering
\small
\setlength{\tabcolsep}{4.5pt}
\caption{Operator structure ablation under matched training. 
Results are mean $\pm$ std over three seeds.}
\vspace{-3mm}
\label{tab:mechanism_ablation}
\resizebox{\linewidth}{!}{
\begin{tabular}{lcccc}
\toprule
Variant
& Scan2Cap C@0.5
& ScanRefer Acc@0.25
& EmbodiedScan 4F F1@0.25
& EmbodiedScan 6F F1@0.25 \\
\midrule
Additive Only
& $80.07{\pm}0.12$
& $57.62{\pm}0.18$
& $41.18{\pm}0.24$
& $40.05{\pm}0.13$ \\
Visual-Conditioned Routing
& $80.71{\pm}0.18$
& $59.56{\pm}0.23$
& $42.48{\pm}0.19$
& $41.69{\pm}0.22$ \\
Generic Cross-Attention
& $80.62{\pm}0.17$
& $59.08{\pm}0.24$
& $42.36{\pm}0.18$
& $41.48{\pm}0.21$ \\
Learnable $3{\times}3$ Generator
& $80.11{\pm}0.14$
& $57.91{\pm}0.20$
& $41.56{\pm}0.15$
& $40.71{\pm}0.18$ \\
Spectral-Controlled $3{\times}3$ Generator
& $80.76{\pm}0.15$
& $59.34{\pm}0.22$
& $42.61{\pm}0.16$
& $41.72{\pm}0.20$ \\
Rational-Controlled $3{\times}3$ Generator
& $81.48{\pm}0.16$
& $61.27{\pm}0.25$
& $43.83{\pm}0.17$
& $42.76{\pm}0.21$ \\
$\mathfrak{so}(3)$--Pad\'e (Ours)
& $\mathbf{82.69{\pm}0.15}$
& $\mathbf{64.56{\pm}0.31}$
& $\mathbf{45.76{\pm}0.13}$
& $\mathbf{44.10{\pm}0.26}$ \\
\bottomrule
\end{tabular}}
\vspace{-5mm}
\end{table*}

\begin{table}[!t]
\begin{minipage}[!t]{0.508\linewidth}
\centering
\caption{
Comparison of numerical realizations for the proposed Lie Group Routing operator.
}
\renewcommand\arraystretch{1}
\resizebox{\linewidth}{!}{
\begin{tabular}{c|cc|ccc}
\toprule
\makecell[c]{\multirow{2}{*}{Method}} & \multicolumn{2}{c|}{Scan2Cap} & \multicolumn{3}{c}{3D Video Detection} \\ 
\cmidrule(l){2-3} \cmidrule(l){4-6}
& C@0.5 & B-4@0.5 & P$_{25}$ & R$_{25}$ & F1$_{25}$ \\
\hline
Rodrigues      & 81.7 & 41.4 & 47.8 & 42.7 & 44.5 \\
Taylor-1       & 81.2 & 41.3 & 47.2 & 42.3 & 43.9 \\
Taylor-2       & 81.6 & 41.5 & 47.6 & 42.8 & 44.2 \\
Taylor-3       & 81.9 & 41.4 & 47.9 & 42.9 & 44.6 \\
Pad\'e-2/2     & 82.4 & 41.6 & 48.4 & 43.3 & 45.3 \\
Pad\'e-1/1 (Ours) & \textbf{82.7} & \textbf{41.7} & \textbf{48.6} & \textbf{44.1} & \textbf{45.8} \\
\bottomrule
\end{tabular}
}
\label{table:ablation1}
\end{minipage}
\hfill
\begin{minipage}[!t]{0.46\linewidth}
\centering
\caption{
Sensitivity to the initial routing bias $b_{\gamma}$ in Lie State Elicitation.
}
\renewcommand\arraystretch{1}
\resizebox{\linewidth}{!}{
\begin{tabular}{c|cc|ccc}
\toprule
\makecell[c]{\multirow{2}{*}{$b_{\gamma}$}} & \multicolumn{2}{c|}{Scan2Cap} & \multicolumn{3}{c}{3D Video Detection} \\ 
\cmidrule(l){2-3} \cmidrule(l){4-6}
& C@0.5 & B-4@0.5 & P$_{25}$ & R$_{25}$ & F1$_{25}$ \\
\hline
$0$  & 81.8 & 41.4 & 47.7 & 43.7 & 45.2 \\
$-1$ & 82.0 & 41.5 & 47.9 & 43.9 & 45.4 \\
$-3$ & 82.5 & 41.7 & 48.4 & \textbf{44.2} & 45.7 \\
$-4$ & \textbf{82.7} & \textbf{41.8} & \textbf{48.6} & 44.1 & \textbf{45.8} \\
$-6$ & 82.3 & 41.6 & 48.2 & 44.0 & 45.6 \\
\bottomrule
\end{tabular}
}
\label{table:ablation2}
\end{minipage}
\vspace{-5mm}
\end{table}

\vspace{-2mm}
\section{Conclusion}
\vspace{-3.5mm}

We presented LieVLM, a Lie routing framework for view-robust 3D MLLMs. Rather than treating geometry as auxiliary context alone, LieVLM uses geometry tokens to predict local $\mathfrak{so}(3)$ states and routing coefficients that govern structured transformations of pretrained visual semantics across viewpoints. Through Lie-State Elicitation, Lie-Group Routing, and Geometry-Aware Routed Fusion, the model injects gated corrections while preserving the original visual-semantic, thereby mitigating viewpoint-induced semantic drift. Experiments across 3D scene understanding and spatial reasoning show consistent gains over additive geometry fusion, with benefits that generalize across benchmarks, training scales, and stronger backbones. Controlled ablations and viewpoint analyses further connect these gains to geometry-conditioned routing and structured transformation design. 


\paragraph{Acknowledgements}
This work is supported by Zhejiang Leading Innovative and Entrepreneur
Team Introduction Program (2024R01007). This work is also supported by
The “Pioneer” and “Leading Goose” Research and Development Program
of Zhejiang (2025C02077). Jingjun Yi sincerely thanks Prof.~Pascal Mettes (University of Amsterdam) for his extensive help in revising the manuscript, greatly improving its clarity and presentation.

{
\small

\bibliographystyle{plainnat}
\bibliography{ref}

}


\appendix

\newpage

\addtocontents{toc}{\protect\setcounter{tocdepth}{2}}
\setcounter{tocdepth}{-1}
\newpage
\begin{center}
    {\LARGE \textbf{Supplementary Material}}
    \vspace{2em}
\end{center}
\tableofcontents
\clearpage

\section{Background: Exponential Map on \texorpdfstring{$\mathrm{SO}(3)$}{SO(3)}}
\label{app:so3_background}

This section provides background on the numerical realization of the exponential action on $\mathrm{SO}(3)$. 
We first recall the exact Rodrigues formula, and then review Taylor and Pad\'e approximations, which are commonly used to implement matrix exponentials in practice.

\subsection{Rodrigues Formula}
\label{app:rodrigues_formula}

Let $\Omega=\omega^\wedge\in\mathfrak{so}(3)$ and $\theta=\|\omega\|_2$.
The exponential map from $\mathfrak{so}(3)$ to $\mathrm{SO}(3)$ admits the classical Rodrigues formula,
\begin{equation}
    \exp(\Omega)=I+A(\theta)\Omega+B(\theta)\Omega^2,
    \label{eq:app_rodrigues}
\end{equation}
where, for $\theta>0$, the scalar coefficients are
\begin{equation}
    A(\theta)=\frac{\sin\theta}{\theta},
    \qquad
    B(\theta)=\frac{1-\cos\theta}{\theta^2}.
    \label{eq:app_rodrigues_coeff}
\end{equation}
For a vector $v\in\mathbb{R}^3$, the corresponding action is
\begin{equation}
    \exp(\Omega)v=v+A(\theta)\Omega v+B(\theta)\Omega^2v.
    \label{eq:app_rodrigues_action}
\end{equation}
The Rodrigues formula is exact in real arithmetic, but direct evaluation of its coefficient ratios requires care near $\theta=0$. Both ratios have removable singularities with limits
\begin{equation}
    \lim_{\theta\rightarrow0}A(\theta)=1,
    \qquad
    \lim_{\theta\rightarrow0}B(\theta)=\frac{1}{2}.
    \label{eq:app_rodrigues_limits}
\end{equation}
We set $A(0)=1$ and $B(0)=1/2$ by continuous extension. Taylor expansions provide regular local approximations for evaluating the coefficients near zero. The apparent singularities therefore belong to the ratio expressions, while the exponential map itself remains smooth.

\subsection{Taylor Expansion}
\label{app:taylor_expansion}

Taylor expansion~\citep{higham2008functions} provides the most direct power-series realization of the matrix exponential. It follows from the analytic definition of the exponential map, which represents $\exp(\Omega)$ as an infinite sum of weighted powers of the generator $\Omega$. 
Truncating this series yields a polynomial approximation whose accuracy improves near the identity as more orders are retained. 
For $\Omega\in\mathfrak{so}(3)$ and $v\in\mathbb{R}^3$, the exponential action can be written as
\begin{equation}
    \exp(\Omega)v=\sum_{k=0}^{\infty}\frac{1}{k!}\Omega^k v
    =v+\Omega v+\frac{1}{2}\Omega^2v+\frac{1}{6}\Omega^3v+\cdots .
    \label{eq:app_taylor_series}
\end{equation}
A finite Taylor approximation of order $K$ is therefore
\begin{equation}
    \Phi_{\mathrm{Taylor}\text{-}K}(\Omega,v)=\sum_{k=0}^{K}\frac{1}{k!}\Omega^k v .
    \label{eq:app_taylor_k}
\end{equation}
This polynomial form is smooth around the identity and avoids the explicit trigonometric ratios in Rodrigues formula, making it suitable for small-angle implementations. 
However, finite Taylor truncations control the response only through the truncation order and $\|\Omega\|$, without the rational normalization introduced by Pad\'e approximations.

\subsection{Pad\'e Approximation}
\label{app:pade_approximation}

Pad\'e approximation~\citep{higham2008functions} provides a rational-function realization of a given analytic function. 
Given a scalar function $f(x)$, its $(m,n)$ Pad\'e approximant takes the form
\begin{equation}
    R_{m,n}(x)=\frac{P_m(x)}{Q_n(x)}
    =\frac{a_0+a_1x+\cdots+a_mx^m}{1+b_1x+\cdots+b_nx^n},
    \label{eq:app_pade_general}
\end{equation}
where the coefficients are chosen such that the Taylor expansion of $R_{m,n}(x)$ matches that of $f(x)$ up to the highest possible order around the expansion point. 
Compared with Taylor truncation, Pad\'e approximation uses both numerator and denominator polynomials, and therefore provides a rational rather than purely polynomial approximation.

For the matrix exponential, Pad\'e approximation replaces the infinite exponential series with a rational matrix function. 
The commonly used diagonal Pad\'e forms approximate $\exp(\Omega)$ by a ratio of two matrix polynomials. 
In particular, the Pad\'e-$1/1$ form is
\begin{equation}
    R_{1,1}(\Omega)=\left(I-\frac{1}{2}\Omega\right)^{-1}\left(I+\frac{1}{2}\Omega\right),
    \label{eq:app_pade_11}
\end{equation}
where $(\cdot)^{-1}$ denotes the matrix inverse, usually implemented by solving a linear system. 
This form is exactly the Cayley map applied to $\Omega/2$. A higher-order diagonal variant is the Pad\'e-$2/2$ form:
\begin{equation}
    R_{2,2}(\Omega)
    =
    \left(I-\frac{1}{2}\Omega+\frac{1}{12}\Omega^2\right)^{-1}
    \left(I+\frac{1}{2}\Omega+\frac{1}{12}\Omega^2\right).
    \label{eq:app_pade_22}
\end{equation}
Both Pad\'e-$1/1$ and Pad\'e-$2/2$ are rational approximations to $\exp(\Omega)$. 
When $\Omega\in\mathfrak{so}(3)$ is skew-symmetric, these diagonal Pad\'e forms preserve the rotation structure because the denominator polynomial is the transpose of the numerator polynomial. 
Thus, they define valid $\mathrm{SO}(3)$ actions while approximating the exact exponential map.

\section{Theoretical Analysis}
\label{app:theoretical_analysis}

This section analyzes the local sensitivity and structural properties of the operator realizations. We examine direct Rodrigues evaluation near the identity, establish Jacobian bounds for Pad\'e actions, and compare the rotation structure and response magnitude of finite Taylor and Pad\'e forms.

\subsection{Direct Rodrigues Evaluation Near the Identity}
\label{app:rodrigues_instability}

Rodrigues formula expresses the exact exponential action through the coefficients $A(\theta)$ and $B(\theta)$. Their direct derivative expressions involve subtracting nearly equal quantities near $\theta=0$, which can amplify rounding errors during evaluation. We examine this computational form while retaining the continuous limits of the underlying functions.

Specifically, differentiating $A(\theta)=\frac{\sin\theta}{\theta}, B(\theta)=\frac{1-\cos\theta}{\theta^2}$ gives
\begin{equation}
    dA=\frac{\theta\cos\theta-\sin\theta}{\theta^3}(\omega^\top d\omega),\quad
    dB=\frac{\theta\sin\theta-2(1-\cos\theta)}{\theta^4}(\omega^\top d\omega).
    \label{eq:app_rodrigues_backward_terms}
\end{equation}
The scalar coefficients in Eq.~\ref{eq:app_rodrigues_backward_terms} have well-defined limits:
\begin{equation}
    \lim_{\theta\to0}\frac{\theta\cos\theta-\sin\theta}{\theta^3}=-\frac{1}{3},\quad
    \lim_{\theta\to0}\frac{\theta\sin\theta-2(1-\cos\theta)}{\theta^4}=-\frac{1}{12}.
\end{equation}

\subsection{Local Jacobian Bounds for Pad\'e Routing}
\label{app:pade_stability}

We establish a local Jacobian bound for the full Pad\'e 1/1 action underlying the operator correction. The derivative is taken with respect to the predicted operator state while holding the visual feature fixed. The argument extends to general Pad\'e forms under a uniform nonsingularity condition on the denominator.

\begin{proposition}[Local Jacobian bound]
\label{prop:pade_stability}
Let $\Omega=\omega^\wedge\in\mathfrak{so}(3)$ and define the Pad\'e-$1/1$ routing action as
\begin{equation}
    \Phi_{1,1}(\omega,v)=v+\frac{\Omega v+\frac{1}{2}\Omega^2v}{1+\frac{1}{4}\|\omega\|_2^2}.
\end{equation}
For any $\rho>0$, there exists a constant $C_\rho<\infty$ such that, for all $\|\omega\|_2\le\rho$ and $v\in\mathbb{R}^3$,
\begin{equation}
    \left\|\frac{\partial \Phi_{1,1}(\omega,v)}{\partial \omega}\right\|\le C_\rho\|v\|_2 .
\end{equation}
\end{proposition}


The Jacobian norm is the operator norm induced by the Euclidean norm.

\begin{proof}
Let
\begin{equation}
    q(\omega)=1+\frac{1}{4}\|\omega\|_2^2,\quad N(\omega,v)=\Omega v+\frac{1}{2}\Omega^2v .
\end{equation}
Then $\Phi_{1,1}(\omega,v)=v+N(\omega,v)/q(\omega)$. 
For any perturbation direction $h\in\mathbb{R}^3$, since $\Omega=\omega^\wedge$ depends linearly on $\omega$, we have
\begin{equation}
    D\Omega[h]=h^\wedge.
\end{equation}
Therefore, the differentials of the numerator and denominator are
\begin{equation}
\label{eq:app_pade_derivatives}
\begin{aligned}
    DN(\omega,v)[h]
    &= h^\wedge v+\frac{1}{2}\left(h^\wedge \Omega v+\Omega h^\wedge v\right),\\
    Dq(\omega)[h]
    &= \frac{1}{2}\omega^\top h .
\end{aligned}
\end{equation}
Since $\|x^\wedge \|_2=\|x\|_2$ for any $x\in\mathbb{R}^3$, when $\|\omega\|_2\le\rho$, we have
\begin{equation}
\label{eq:app_pade_bounds}
\begin{aligned}
    \|N(\omega,v)\|_2
    &\le \left(\rho+\frac{1}{2}\rho^2\right)\|v\|_2,\\
    \|DN(\omega,v)[h]\|_2
    &\le (1+\rho)\|h\|_2\|v\|_2,\\
    q(\omega)&\ge 1,\quad
    |Dq(\omega)[h]|\le\frac{1}{2}\rho\|h\|_2 .
\end{aligned}
\end{equation}

Using the above relations, the directional derivative of $\Phi_{1,1}$ can be bounded as follows:
\begin{equation}
\label{eq:app_pade_main_bound}
\begin{aligned}
    \|D\Phi_{1,1}(\omega,v)[h]\|_2
    &\overset{(1)}{=}
    \left\|
    \frac{DN(\omega,v)[h]q(\omega)-N(\omega,v)Dq(\omega)[h]}{q(\omega)^2}
    \right\|_2\\
    &\overset{(2)}{\le}
    \frac{\|DN(\omega,v)[h]\|_2}{q(\omega)}
    +
    \frac{\|N(\omega,v)\|_2|Dq(\omega)[h]|}{q(\omega)^2}\\
    &\overset{(3)}{\le}
    (1+\rho)\|h\|_2\|v\|_2
    +
    \frac{1}{2}\rho\left(\rho+\frac{1}{2}\rho^2\right)\|h\|_2\|v\|_2\\
    &=
    \left((1+\rho)+\frac{1}{2}\rho\left(\rho+\frac{1}{2}\rho^2\right)\right)\|h\|_2\|v\|_2\\
    &=
    C_\rho\|h\|_2\|v\|_2 ,
\end{aligned}
\end{equation}
where $C_\rho=(1+\rho)+\frac{1}{2}\rho(\rho+\frac{1}{2}\rho^2)<\infty$.

The derivation of Eq.~\ref{eq:app_pade_main_bound} follows.

(1) follows from the quotient rule applied to $\Phi_{1,1}(\omega,v)=v+N(\omega,v)/q(\omega)$.

(2) follows from the triangle inequality and norm sub-multiplicativity.

(3) follows from the bounds in Eq.~\ref{eq:app_pade_bounds} and the fact that $q(\omega)\ge1$.

Taking the supremum over all perturbation directions with $\|h\|_2=1$ gives
\begin{equation}
    \left\|\frac{\partial \Phi_{1,1}(\omega,v)}{\partial \omega}\right\|
    \le C_\rho\|v\|_2 .
\end{equation}
This completes the proof.
\end{proof}

\begin{remark}[Extension to Pad\'e-$m/n$]
The above proof uses Pad\'e-$1/1$ as the default case in our main model, but the same argument applies to a general Pad\'e-$m/n$ routing action under a local nonsingularity condition. 
Let
\begin{equation}
    P_m(\Omega)=\sum_{i=0}^{m}p_i\Omega^i,\quad
    Q_n(\Omega)=\sum_{j=0}^{n}q_j\Omega^j,\quad q_0=1,
\end{equation}
and define
\begin{equation}
    \Phi_{m,n}(\omega,v)=Q_n(\Omega)^{-1}P_m(\Omega)v .
\end{equation}
Assume that, for all $\|\omega\|_2\le\rho$, the denominator matrix is locally nonsingular:
\begin{equation}
    \sigma_{\min}(Q_n(\Omega))\ge \alpha_{m,n,\rho}>0 .
\end{equation}
We further define the coefficient-dependent quantities
\begin{equation}
    P_{m,\rho}=\sum_{i=0}^{m}|p_i|\rho^i,\quad
    P'_{m,\rho}=\sum_{i=1}^{m}i|p_i|\rho^{i-1},\quad
    Q'_{n,\rho}=\sum_{j=1}^{n}j|q_j|\rho^{j-1}.
\end{equation}

For any perturbation direction $h\in\mathbb{R}^3$, since $D\Omega[h]=h^\wedge$ and $\|h^\wedge\|_2=\|h\|_2$, we have
\begin{equation}
    \|D\Omega^k[h]\|_2
    =
    \left\|\sum_{\ell=0}^{k-1}\Omega^\ell h^\wedge \Omega^{k-1-\ell}\right\|_2
    \le k\rho^{k-1}\|h\|_2 .
    \label{eq:app_D_power_bound}
\end{equation}
Therefore,
\begin{equation}
\label{eq:app_general_pade_basic_bounds}
\begin{aligned}
    \|P_m(\Omega)v\|_2 &\le P_{m,\rho}\|v\|_2,\\
    \|DP_m(\Omega)[h]v\|_2 &\le P'_{m,\rho}\|h\|_2\|v\|_2,\\
    \|DQ_n(\Omega)[h]\|_2 &\le Q'_{n,\rho}\|h\|_2,\\
    \|Q_n(\Omega)^{-1}\|_2 &\le \alpha_{m,n,\rho}^{-1}.
\end{aligned}
\end{equation}

Using these bounds, the directional derivative of $\Phi_{m,n}$ can be bounded as
\begin{equation}
\label{eq:app_general_pade_main_bound}
\begin{aligned}
    \|D\Phi_{m,n}(\omega,v)[h]\|_2
    &\overset{(1)}{=}
    \left\|
    -Q_n^{-1}(DQ_n[h])Q_n^{-1}P_m v
    +
    Q_n^{-1}(DP_m[h])v
    \right\|_2\\
    &\overset{(2)}{\le}
    \|Q_n^{-1}\|_2\|DP_m[h]v\|_2
    +
    \|Q_n^{-1}\|_2^2\|DQ_n[h]\|_2\|P_m v\|_2\\
    &\overset{(3)}{\le}
    \left(
    \frac{P'_{m,\rho}}{\alpha_{m,n,\rho}}
    +
    \frac{Q'_{n,\rho}P_{m,\rho}}{\alpha_{m,n,\rho}^{2}}
    \right)\|h\|_2\|v\|_2\\
    &= C_{m,n,\rho}\|h\|_2\|v\|_2 ,
\end{aligned}
\end{equation}
where
\begin{equation}
    C_{m,n,\rho}
    =
    \frac{P'_{m,\rho}}{\alpha_{m,n,\rho}}
    +
    \frac{Q'_{n,\rho}P_{m,\rho}}{\alpha_{m,n,\rho}^{2}} .
\end{equation}

The derivation of Eq.~\ref{eq:app_general_pade_main_bound} follows.

(1) follows from the product rule and the inverse differential identity $D(Q^{-1})=-Q^{-1}(DQ)Q^{-1}$.

(2) follows from the triangle inequality and norm sub-multiplicativity.

(3) follows from the bounds in Eq.~\ref{eq:app_general_pade_basic_bounds}.

Taking the supremum over all perturbation directions with $\|h\|_2=1$ gives
\begin{equation}
    \left\|\frac{\partial \Phi_{m,n}(\omega,v)}{\partial \omega}\right\|
    \le C_{m,n,\rho}\|v\|_2 .
\end{equation}
Thus, the Pad\'e action has a locally bounded Jacobian under the stated denominator condition. For Pad\'e 1/1, the correction used in the main model has the strictly positive scalar denominator $1+\|\omega\|_2^2/4$.
\end{remark}

\begin{remark}
The bound characterizes the sensitivity of the exact Pad\'e action to the predicted operator state. The implemented triplet correction is $c^{\mathrm{pade}}_{n,m}=\Phi_{1,1}(\omega_{n,m},v_{n,m})-v_{n,m}$. For fixed $v$, subtracting $v$ leaves the Jacobian with respect to $\omega$ unchanged, so the same bound applies to the implemented correction. This result concerns input sensitivity in exact arithmetic, independently of finite precision rounding error.
\end{remark}

\subsection{Structure and Response of Taylor and Pad\'e Actions}
\label{app:taylor_vs_pade}

Finite Taylor and diagonal Pad\'e forms provide smooth realizations of the operator action near the identity. We compare their rotation structure and the magnitude of the induced semantic change.

First, finite Taylor truncations do not generally preserve the $\mathrm{SO}(3)$ constraint. 
Let $T_K(\Omega)=\sum_{k=0}^{K}\frac{1}{k!}\Omega^k$. 
Since $\Omega^\top=-\Omega$, its orthogonality defect is
\begin{equation}
    T_K(\Omega)^\top T_K(\Omega)-I=T_K(-\Omega)T_K(\Omega)-I .
\end{equation}
This defect is composed of higher-order powers of $\Omega$. 
For example,
\begin{equation}
\begin{aligned}
    T_1(\Omega)^\top T_1(\Omega)-I
    &=-\Omega^2,\\
    T_2(\Omega)^\top T_2(\Omega)-I
    &=\frac{1}{4}\Omega^4,\\
    T_3(\Omega)^\top T_3(\Omega)-I
    &=-\frac{1}{12}\Omega^4-\frac{1}{36}\Omega^6 .
\end{aligned}
\label{eq:app_taylor_orth_defect}
\end{equation}
These expressions quantify the orthogonality defect of finite Taylor truncations. The defect depends on both the truncation order and generator magnitude, and finite truncation does not guarantee a rotation matrix for arbitrary generators.
As $\|\Omega\|_2=\|\omega\|_2$ increases, these residual high-order terms become more visible, so Taylor truncations do not strictly define rotation matrices unless additional projection or re-orthogonalization is applied.

In contrast, diagonal Pad\'e forms preserve the orthogonal structure for skew-symmetric generators. 
A diagonal Pad\'e approximation can be written as
\begin{equation}
    R_{\ell,\ell}(\Omega)=Q_\ell(\Omega)^{-1}P_\ell(\Omega),\quad Q_\ell(\Omega)=P_\ell(-\Omega).
\end{equation}
Since $\Omega^\top=-\Omega$, we have $Q_\ell(\Omega)=P_\ell(\Omega)^\top$. 
Therefore,
\begin{equation}
    R_{\ell,\ell}(\Omega)^\top R_{\ell,\ell}(\Omega)
    =
    Q_\ell(\Omega)P_\ell(\Omega)^{-1}Q_\ell(\Omega)^{-1}P_\ell(\Omega)
    =
    I,
\end{equation}
where the last equality follows because $P_\ell(\Omega)$ and $Q_\ell(\Omega)$ are polynomials of the same matrix and hence commute. 
If $Q_\ell(t\Omega)$ remains nonsingular for all $t\in[0,1]$, continuity from $R_{\ell,\ell}(0)=I$ further gives $\det R_{\ell,\ell}(\Omega)=1$.
In particular, the Pad\'e 1/1 and Pad\'e 2/2 forms used here have nonsingular denominators for real skew-symmetric generators and therefore yield matrices in $\mathrm{SO}(3)$.

Second, Taylor truncations are purely polynomial. 
For the finite Taylor approximation in Eq.~\ref{eq:app_taylor_k}, the operator induced change satisfies
\begin{equation}
    \|\Phi_{\mathrm{Taylor}\text{-}K}(\Omega,v)-v\|_2
    \le
    \left(\sum_{k=1}^{K}\frac{\|\omega\|_2^k}{k!}\right)\|v\|_2 .
    \label{eq:app_taylor_bound}
\end{equation}
By contrast, the Pad\'e-$1/1$ routing form has the bound
\begin{equation}
    \|\Phi_{1,1}(\omega,v)-v\|_2
    \le
    \frac{\|\omega\|_2+\frac{1}{2}\|\omega\|_2^2}{1+\frac{1}{4}\|\omega\|_2^2}\|v\|_2 .
    \label{eq:app_pade_bound}
\end{equation}
The denominator in Eq.~\ref{eq:app_pade_bound} provides a rational normalization of the linear and quadratic generator actions. 
Finite Taylor forms avoid direct evaluation of the Rodrigues coefficient ratios, while Pad\'e 1/1 additionally preserves the rotation structure and introduces rational normalization of the operator response.

\section{Experimental Setting and Details}
\subsection{Implementation Details}
\label{app:imp}
LieVLM uses VGGT-1B as the geometry encoder and Qwen2.5-VL-7B as the default language backbone. For the scaled spatial reasoning experiments in Sec.~\ref{sec:general_spatial_reasoning}, we additionally instantiate matched VG LLM and LieVLM variants with Qwen3-VL-8B-Instruct. Within each matched setting, the two models use identical initialization, training data, optimization, and evaluation, and differ only in additive geometry fusion or Lie routing. For the original setting, we train for one epoch using Adam with a batch size of 64.
The learning rate is warmed up with a ratio of 0.03 to a peak value of $1\times10^{-5}$, and then linearly decayed to zero. At each iteration, one data source is sampled from the mixed training corpus to form the batch.
During training, the 2D visual encoder, the 3D geometry encoder, and the multimodal connector are kept frozen. We optimize the language backbone together with the proposed LieVLM modules, including the rotation head, routing head, and routed fusion layers. 
The output projection is applied to the correction $c^{\mathrm{pade}}_{n,m} =
R_{1,1}(\Omega_{n,m})v_{n,m}-v_{n,m}$,
rather than to the full transformed triplet.
The routing head $W_{\gamma}:\mathbb{R}^{H}\rightarrow\mathbb{R}^{H}$ predicts a token and channel wise gate $\gamma\in(0,1)^{N\times H}$, which modulates the routed feature after the output projection $W_o$.
The learnable routing bias $b_{\gamma}$ is initialized to $-4$ and optimized jointly with the routing head.
For the spectral-controlled $3{\times}3$ generator in Table~\ref{tab:mechanism_ablation}, we retain the prediction head and parameter count of the unconstrained dynamic-matrix baseline but normalize each predicted residual matrix as
\begin{equation}
    \bar G_{n,m}
    =
    \frac{\tau G_{n,m}}{\max\!\left(\tau,\lVert G_{n,m}\rVert_2\right)},
    \qquad
    c^{\mathrm{spec}}_{n,m}=\bar G_{n,m}v_{n,m}.
\end{equation}
We set $\tau=2\sqrt{3/7}\approx1.309$, matching the maximum spectral norm of $R_{1,1}(\Omega)-I$ under the component-wise bounded state $\omega=\tanh(W_{\omega}Z)$. 
The baseline otherwise shares the direct geometry path, routing gate, output projection, initialization, and training protocol with the unconstrained generator.
For the rational-controlled $3{\times}3$ generator, we retain the same unconstrained prediction head but replace the hard spectral cap with
\begin{equation}
    c^{\mathrm{rat}}_{n,m} = \frac{G_{n,m}v_{n,m}}
{1+\frac18\|G_{n,m}\|_F^2}.
\end{equation}
The factor $1/8$ matches the Pad\'e-1/1 denominator for a skew-symmetric generator because $\|\omega^\wedge\|_F^2=2\|\omega\|_2^2$. This variant retains the same direct geometry path, routing gate, output projection, parameter count, and training protocol.
For visual-conditioned routing, we retain the complete $\mathfrak{so}(3)$--Pad\'e routing architecture and direct geometry path, but replace the conditioning input in Eq.~\ref{eq:lie_state_prediction} with $Z^{2d}=\operatorname{LN}(F^{2d})$. Both $\omega$ and $\gamma$ are therefore predicted from visual semantics rather than geometry tokens, with all other components and training settings unchanged.
The MLLM interface width is $H=3584$ for Qwen2.5-VL-7B and $H=4096$ for Qwen3-VL-8B. We set $D_o=3\lfloor H/3\rfloor$, giving $(D_o,M)=(3582,1194)$ and $(4095,1365)$ for the two backbones, respectively. The projections $W_v$ and $W_{\omega}$ map $H$ to $D_o$, $W_{\gamma}$ maps $H$ to $H$, and the bias free $W_o$ maps $D_o$ back to $H$.
The four projection matrices contain $3HD_o+H^2$ weights, corresponding to $51.36$M and $67.10$M parameters for the two settings, excluding biases.
Following the geometry fusion interface of VG LLM~\citep{zheng2025learning}, we adopt the same VGGT-to-MLLM token alignment and projection procedure without modification. The aligned VGGT features are projected to the interface width $H$ before fusion.
The resulting geometry tokens $F^{3d}$ share the token indexing and channel dimension of $F^{2d}$. The same aligned geometry representation supplies both the direct geometry path and Lie State Elicitation, so geometry injection and operator prediction operate on a common token grid.
All experiments are conducted on 8 H100 80G GPUs. For 3D scene understanding, training takes approximately 12 hours for LieVLM. 
For the original 297K VG LLM spatial reasoning setting, training takes approximately 9 hours.

\subsection{Data Preparation}
\label{appx:data_pre}

We use publicly available datasets and follow the VG LLM \citep{zheng2025learning} preprocessing protocol to convert them into the input-output format required by LieVLM.




\textbf{3D Scene Understanding.}
For ScanRefer and Scan2Cap, we sample 32 RGB frames per scene. For 3D video object detection, each sample contains a 4-frame clip at 1 FPS, with annotations in the first-frame coordinate system.

\textbf{SPAR-7M.}
We process SPAR-7M with the official implementation\footnote{\url{https://github.com/fudan-zvg/spar}}, retaining spatial QA samples with valid visual inputs and rendering visual markers using the released pipeline. We exclude navigation samples because of their variable-length image sequences.

\textbf{LLaVA-Video-178K.}
We use the LLaVA-Hound split as general video instruction data, sampling frames at 2 FPS and keeping 4--8 frames per video.

\textbf{297K Spatial Reasoning Setting.}
Following VG LLM~\citep{zheng2025learning}, we use 234K SPAR-7M samples and 63K LLaVA-Hound samples, corresponding to 3\% and 25\% of the two source datasets. The same composition and preprocessing are used for VG LLM and LieVLM.

\subsection{Large Scale Spatial Reasoning Setting}
\label{app:large_scale_spatial}

\paragraph{Training Mixture.}
Following the public data sources and annotations released with CA-MLLM, we construct a 3.05M example generalist mixture covering video instruction following, spatial question answering, 3D grounding and captioning, 3D detection, and object centric spatial reasoning. We retain our own architecture and training pipeline and do not use the camera ray embeddings, explicit camera intrinsics, camera specific augmentation, or UniDepth distillation introduced by CA-MLLM. Table~\ref{tab:large_scale_mixture} summarizes the complete composition.

\begin{table*}[ht]
\centering
\small
\setlength{\tabcolsep}{5pt}
\caption{
Composition of the 3.05M generalist training mixture.
}
\label{tab:large_scale_mixture}
\resizebox{\textwidth}{!}{
\begin{tabular}{llll}
\toprule
\textbf{Data Component}
& \textbf{Public Sources}
& \textbf{Sampling}
& \textbf{Examples} \\
\midrule
General video instruction
& LLaVA Hound split of LLaVA Video 178K
& Full released split
& 0.26M \\

Large scale spatial QA
& SPAR subsets from ScanNet, ScanNet++, Structured3D, and RxR
& 10\% of each training subset
& 1.63M \\

3D grounding and captioning
& ScanRefer and Scan2Cap
& Official training splits
& 0.07M \\

Other spatially grounded tasks
& ScanNet, 3RScan, Matterport3D, ARKitScenes, Objectron, and SUN RGB-D
& Released annotations
& 1.09M \\
\midrule
\textbf{Total}
& --
& --
& \textbf{3.05M} \\
\bottomrule
\end{tabular}}
\end{table*}

\paragraph{Matched Training and Evaluation.}
The Qwen2.5 variants start from the same Qwen2.5-VL-7B and VGGT-1B checkpoints, while the Qwen3 variants start from the same Qwen3-VL-8B-Instruct and VGGT-1B checkpoints. Within each pair, VG LLM and LieVLM use identical training examples, training settings, and evaluation implementations. The only architectural difference is additive geometry fusion or Lie routing. The evaluated checkpoints receive no benchmark specific fine tuning.

\paragraph{Data Exposure.}
The complete pretraining and post training data of Qwen2.5-VL are not publicly auditable, so prior exposure to ScanNet or ScanNet++ cannot be excluded. Such exposure may affect the absolute scores of both variants. Within each matched comparison, VG LLM and LieVLM use the same base checkpoint and instruction tuning data, so this uncertainty does not create a differential data exposure advantage for LieVLM. CV-Bench-3D and BLINK-Spatial further provide evaluation outside the original ScanNet centered benchmark suite.

\subsection{Evaluation Details}
Following VG LLM \citep{zheng2025learning}, we evaluate LieVLM with LMMs-Eval\footnote{\url{https://github.com/EvolvingLMMs-Lab/lmms-eval}} and use customized task interfaces when necessary. During inference, all responses are generated with greedy decoding.

\textbf{3D Dense Captioning.}
For Scan2Cap, we follow the common two-stage evaluation protocol~\citep{zhu2024llava,zheng2025video}. Object proposals are first obtained from the Mask3D detections provided by LEO~\citep{huang2023embodied}. We transform each proposal center into the reference-frame coordinate system using the corresponding extrinsic matrices~\citep{wang2024embodiedscan}, and then prompt the model to describe the object conditioned on this coordinate. Generated captions are evaluated with CIDEr, BLEU-4, ROUGE, and METEOR using the COCO caption evaluation toolkit\footnote{\url{https://github.com/tylin/coco-caption}}.

\textbf{3D Visual Grounding.}
For ScanRefer, the model predicts both the frame where the referred object is visible and the corresponding oriented 3D bounding box in that frame's camera coordinates. The predicted box is transformed into the world coordinate system before evaluation. Since the original ground-truth boxes are axis-aligned, we append zero rotation angles to obtain oriented boxes for IoU computation. We then report Acc@0.25 and Acc@0.5. 
For the viewpoint trend in Figure~\ref{fig:teaser}(b), we retain the corresponding raw IoU before thresholding to obtain a continuous per-sample measure.
For proposal refinement, we compare the predicted box with Mask3D proposals from LEO~\citep{huang2023embodied} and select the proposal with the highest IoU.

\textbf{3D Video Object Detection.}
For each predicted box, we perform category-wise greedy matching against unused ground-truth boxes. A prediction is counted as a true positive if its best unmatched ground-truth box of the same category exceeds the IoU threshold of 0.25. Matched ground-truth boxes are removed from later matching. Precision, recall, and F1 are then computed for each category and averaged over the 20 common classes.

\textbf{Spatial Reasoning.}
For VSI-Bench~\citep{yang2025thinking}, we use the default LMMs-Eval configuration. Each video is uniformly sampled into 32 frames. Numerical-answer tasks are evaluated with Mean Relative Accuracy, while multiple-choice tasks are evaluated with standard accuracy.

\section{Additional Experimental Results}
\label{app:additional_results}

\subsection{Controlled Viewpoint Responses}
\label{app:viewpoint_responses}

\begin{figure*}[ht]
    \centering
    \includegraphics[width=0.96\textwidth]{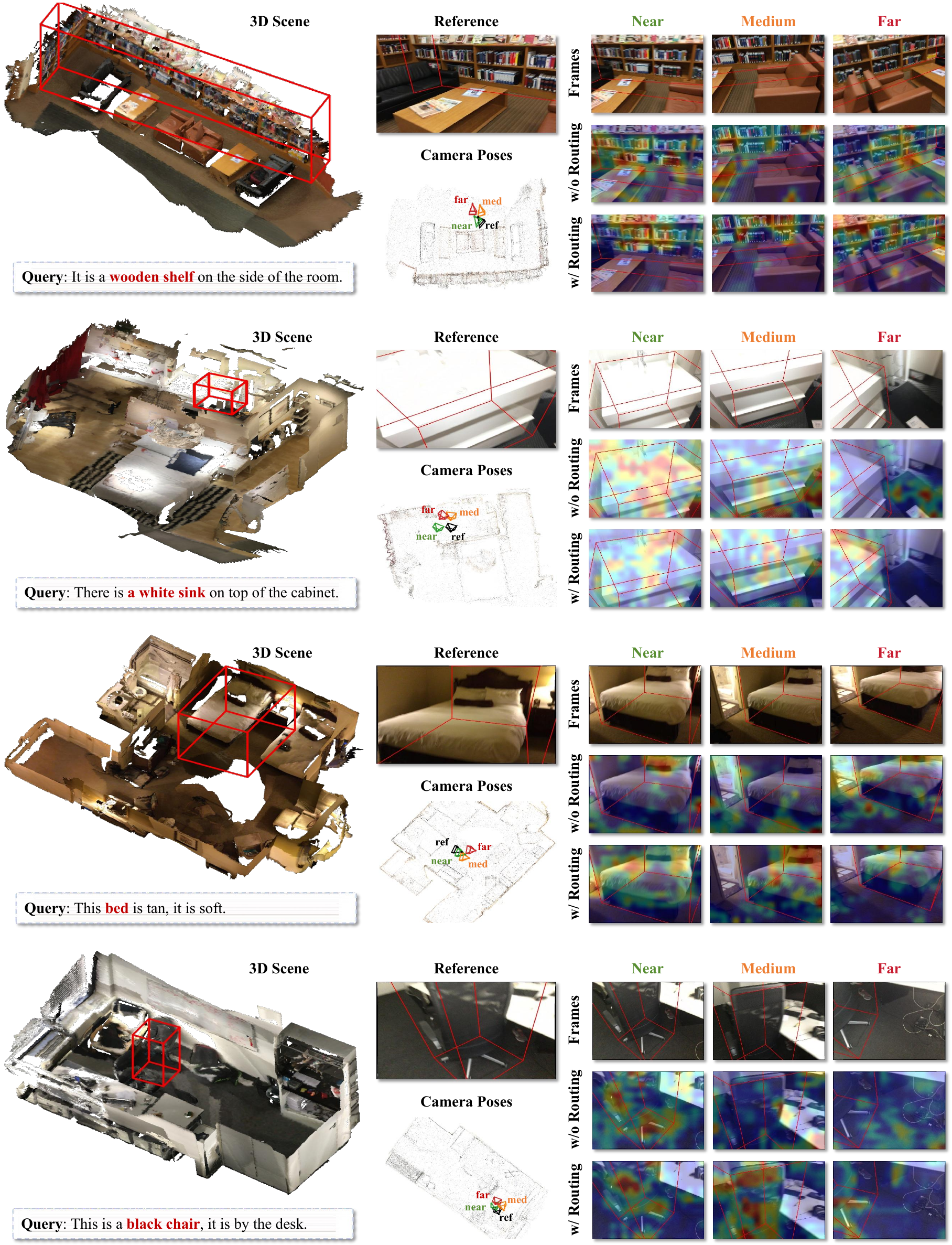}
        \caption{
    Additional controlled viewpoint response examples extending Fig.~\ref{fig:teaser}(a).
    query-relevant activations are shown across reference, near, medium, and far views without and with Lie routing.
    Warmer regions indicate stronger support for the target phrase.
    Red wireframes denote projected ground truth boxes used only for visualization and PBO@0.2.
    }
    \label{fig:viewpoint_responses}
    \vspace{-3mm}
\end{figure*}

Figure~\ref{fig:viewpoint_responses} extends the controlled analysis in Fig.~\ref{fig:teaser}a with four additional scene target query examples. Across the wooden shelf, white sink, tan bed, and black chair cases, 
responses without routing vary substantially across viewpoints and often become diffuse or shift away from the target.
Lie routing consistently preserves more localized target responses across the near, medium, and far views, supporting the same cross view behavior observed in the main teaser.

\paragraph{Activation Localization.}
Following the controlled setting in Fig.~\ref{fig:teaser}(a), each example fixes the scene, target, query, checkpoint, and reference view, chosen as the first frame containing the target, and varies only the viewpoint along the same camera trajectory.
Near, medium, and far views are selected by increasing reference relative rotation distance $d_{\mathrm{rot}}(r,v)$. Let $H_v(p)$ denote the final decoder feature at visual token $p$, and let $\mathcal Q_{\mathrm{t}}$ contain the target phrase tokens. The query-relevant activation is
\begin{equation}
    A_v(p)
    =
    \left|\mathcal Q_{\mathrm{t}}\right|^{-1}
    \sum_{c\in\mathcal Q_{\mathrm{t}}}
    \left[W_c^{\top}H_v(p)\right]_{+},
\end{equation}
where $W_c$ is the vocabulary projection for token $c$, and $[x]_{+}=\max(x,0)$. 
Let $C_v$ denote the top 20 percent activation locations and let $B_v$ denote the projected target region. We measure target localization using
\begin{equation}
    \mathrm{PBO@0.2}=100\frac{|C_v\cap B_v|}{|C_v|}.
\end{equation}

PBO quantifies target localization within each view rather than monotonic degradation with viewpoint distance. The projected target boxes are used only for visualization and evaluation, and are not provided to the model.
Lie routing improves PBO@0.2 in all 15 evaluated views and increases the overall mean from $39.2$ to $71.9$. The improvements remain consistent across the near, medium, and far viewpoint groups.

\begin{table*}[t]
\centering
\small
\setlength{\tabcolsep}{6pt}
\caption{
Target overlap of query-relevant activations under controlled viewpoint variation.
PBO@0.2 reports the percentage of top response locations inside the projected target box.
}
\label{tab:pbo_controlled}
\begin{tabular}{lllccc}
\toprule
\textbf{Scene}
& \textbf{Target}
& \textbf{Routing}
& \textbf{Near}
& \textbf{Medium}
& \textbf{Far} \\

\midrule
\multirow{2}{*}{\texttt{scene0011\_00}}
& \multirow{2}{*}{Long table}
& Without Routing
& 44.6 & 2.7 & 49.5 \\
&
& Lie Routing
& \textbf{81.6} & \textbf{49.4} & \textbf{62.8} \\

\midrule
\multirow{2}{*}{\texttt{scene0064\_00}}
& \multirow{2}{*}{Wooden shelf}
& Without Routing
& 82.6 & 64.7 & 34.0 \\
&
& Lie Routing
& \textbf{86.2} & \textbf{74.0} & \textbf{86.7} \\

\midrule
\multirow{2}{*}{\texttt{scene0246\_00}}
& \multirow{2}{*}{White sink}
& Without Routing
& 73.5 & 40.6 & 15.6 \\
&
& Lie Routing
& \textbf{95.1} & \textbf{71.5} & \textbf{80.7} \\

\midrule
\multirow{2}{*}{\texttt{scene0645\_00}}
& \multirow{2}{*}{Tan bed}
& Without Routing
& 48.8 & 11.5 & 30.6 \\
&
& Lie Routing
& \textbf{83.2} & \textbf{28.2} & \textbf{70.6} \\

\midrule
\multirow{2}{*}{\texttt{scene0700\_00}}
& \multirow{2}{*}{Black chair}
& Without Routing
& 48.3 & 31.9 & 8.9 \\
&
& Lie Routing
& \textbf{76.8} & \textbf{77.8} & \textbf{53.8} \\

\midrule
\multirow{2}{*}{Mean}
&
& Without Routing
& 59.6 & 30.3 & 27.7 \\
&
& Lie Routing
& \textbf{84.6} & \textbf{60.2} & \textbf{70.9} \\

\bottomrule
\end{tabular}
\end{table*}

\subsection{3D Video Object Detection on EmbodiedScan}
\label{app:embodiedscan}

\paragraph{Category-Level Results.}
\begin{table*}[!t]
\centering
\caption{The detailed results on 3D video object detection in 4-frame setting. 
We report the F1 score for all categories.}
\resizebox{0.72\linewidth}{!}{
\setlength{\tabcolsep}{3mm}
\begin{tabular}{lccc}
\toprule
Object & \rotatebox{0}{Qwen2.5-VL} & \rotatebox{0}{VG LLM} & \rotatebox{0}{LieVLM (Ours)} \\ 
\midrule
chair & 41.2 & 54.0 & 53.5\\
pillow & 13.2 & 22.9 & 25.7\\
cabinet & 11.6 & 17.1 & 21.3\\
table & 36.5 & 46.5 & 56.8\\
lamp & 9.0 & 18.9 & 23.3\\
couch & 41.1 & 47.0 & 55.1\\
desk & 34.5 & 41.0 & 45.4\\
stand & 33.7 & 46.1 & 50.2\\
bed & 68.2 & 74.1 & 82.9\\
backpack & 33.3 & 41.4 & 49.5\\
bathtub & 36.6 & 42.1 & 75.9\\
ottoman & 0.0 & 0.0 & 0.0\\
dresser & 38.2 & 49.2 & 47.5\\
bin & 30.2 & 39.8 & 43.0\\
toilet & 68.7 & 82.5 & 86.2\\
refrigerator & 38.5 & 53.6 & 60.0\\
stove & 61.2 & 68.6 & 68.5\\
microwave & 21.3 & 33.3 & 24.1\\
monitor & 21.3 & 34.0 & 29.7\\
computer & 12.0 & 11.4 & 16.5\\
\bottomrule

\end{tabular}
}
\label{tab:3d_det_4_frame}
\vspace{4mm}
\centering
\caption{The detailed results on 3D video object detection in 6-frame setting. 
We report the F1 score for all categories.}
\resizebox{0.72\linewidth}{!}{
\setlength{\tabcolsep}{3mm}
\begin{tabular}{lccc}
\toprule
Object & \rotatebox{0}{Qwen2.5-VL} & \rotatebox{0}{VG LLM} & \rotatebox{0}{LieVLM (Ours)} \\ 
\midrule
chair & 36.1 & 48.7 & 50.0\\
pillow & 12.4 & 18.9 & 22.8\\
cabinet & 10.6 & 17.9 & 19.6\\
table & 32.7 & 44.8 & 54.2\\
lamp & 8.0 & 15.9 & 18.7\\
couch & 40.7 & 46.4 & 51.7\\
desk & 34.4 & 32.3 & 45.7\\
stand & 31.3 & 47.3 & 43.1\\
bed & 64.6 & 75.8 & 83.7\\
backpack & 28.6 & 40.4 & 46.3\\
\midrule
bathtub & 38.4 & 40.4 & 79.9 \\
ottoman & 2.7 & 5.3 & 0.1\\
dresser & 36.4 & 44.5 & 46.6\\
bin & 25.0 & 38.5 & 35.5\\
toilet & 68.6 & 83.2 & 86.5\\
refrigerator & 33.2 & 62.0 & 59.6\\
stove & 55.9 & 69.5 & 77.5\\
microwave & 7.8 & 25.8 & 19.6\\
monitor & 20.3 & 31.7 & 25.6\\
computer & 5.2 & 11.6 & 15.3\\
\bottomrule

\end{tabular}
}
\label{tab:3d_det_6_frame}
\vspace{-4mm}
\end{table*}

Tables~\ref{tab:3d_det_4_frame} and~\ref{tab:3d_det_6_frame} provide category-level F1 scores for 3D video object detection. LieVLM improves over Qwen2.5-VL on nearly all categories, confirming the benefit of geometry-aware modeling for cross-frame 3D localization. Compared with VG LLM, LieVLM also shows clear gains on many spatially prominent categories, including table, couch, bed, bathtub, toilet, refrigerator, and computer in the 4-frame setting, as well as chair, table, desk, bed, bathtub, toilet, stove, and computer in the 6-frame setting. These improvements suggest that Lie-Group Routing helps preserve object orientation and spatial consistency when detections are aggregated into the first-frame coordinate system.

We also observe several remaining challenges. For categories with limited visual evidence or ambiguous geometry, such as ottoman, microwave, and monitor, LieVLM does not always improve over VG LLM. In particular, ottoman remains difficult in both frame settings, likely due to sparse instances and weak appearance cues. These results indicate that operatorized geometry substantially improves common object localization, while rare or small objects still require stronger supervision or more reliable visual evidence.

\vspace{-3mm}
\paragraph{Multi-Seed Reliability.}
\label{app:vod_reliability}
We further evaluate the stability of the matched VG LLM and LieVLM comparison over three random seeds. 
Table~\ref{tab:det_multiseed} provides detailed three-seed detection statistics consistent with the F1 results in Table~\ref{tab:mechanism_ablation}. LieVLM improves all six aggregate precision, recall, and F1 metrics in every matched run, with 4-frame and 6-frame F1 gains of $4.58\pm0.11$ and $4.05\pm0.13$, respectively.

\begin{table}[t]
\centering
\caption{Multi-seed reliability on EmbodiedScan. Results are mean $\pm$ std over three matched random seeds.}
\label{tab:det_multiseed}
\resizebox{\linewidth}{!}{
\begin{tabular}{llcccc}
\toprule
Setting & Metric & VG LLM & LieVLM & Paired Gain & Positive Seeds \\
\midrule
4-Frame
& Prec.@0.25  & $43.19\pm0.25$ & $48.26\pm0.14$ & $+5.07\pm0.11$ & 3/3 \\
& Recall@0.25 & $39.44\pm0.23$ & $43.78\pm0.12$ & $+4.34\pm0.11$ & 3/3 \\
& F1@0.25     & $41.18\pm0.24$ & $45.76\pm0.13$ & $+4.58\pm0.11$ & 3/3 \\
\midrule
6-Frame
& Prec.@0.25  & $43.35\pm0.14$ & $47.58\pm0.28$ & $+4.23\pm0.14$ & 3/3 \\
& Recall@0.25 & $38.57\pm0.12$ & $41.60\pm0.24$ & $+3.03\pm0.12$ & 3/3 \\
& F1@0.25     & $40.05\pm0.13$ & $44.10\pm0.26$ & $+4.05\pm0.13$ & 3/3 \\
\bottomrule
\end{tabular}}
\end{table}

\subsection{Detailed VSI-Bench Results}
\label{app:vsi_details}

\paragraph{Full Category Comparison.}
Table~\ref{tab:vsi_full} complements the compact scaling comparison in Table~\ref{tab:vsi_scaling} with all eight VSI-Bench categories and a broader set of prior MLLMs and spatial models. The Qwen3 matched comparison improves seven of eight categories, while the original Qwen2.5 setting provides complementary three seed evidence. Category results are unavailable for the two external Qwen3 scale references, for which only the reported benchmark averages are shown.

\begin{table*}[t]
\centering
\scriptsize
\setlength{\tabcolsep}{3.0pt}
\caption{
Detailed VSI-Bench comparison across model and scaling settings.
Green values denote changes over the corresponding controlled VG LLM baseline.
}
\vspace{-2.5mm}
\label{tab:vsi_full}
\resizebox{0.99\textwidth}{!}{
\begin{tabular}{lccccccccc}
\toprule
\multirow{2}{*}{\textbf{Method}}
& \multirow{2}{*}{\textbf{Avg.}}
& \multicolumn{4}{c}{\textbf{Numerical Answer}}
& \multicolumn{4}{c}{\textbf{Multiple Choice Answer}} \\
\cmidrule(lr){3-6}
\cmidrule(lr){7-10}
&
& \textbf{Obj. Count}
& \textbf{Abs. Dist.}
& \textbf{Obj. Size}
& \textbf{Room Size}
& \textbf{Rel. Dist.}
& \textbf{Rel. Dir.}
& \textbf{Route Plan}
& \textbf{Appr. Order} \\
\midrule

\rowcolor{gray!15}
\multicolumn{10}{l}{\textit{Prior MLLMs and Spatial Models}} \\

GPT-4o
& 34.0
& 46.2
& 5.3
& 43.8
& 38.2
& 37.0
& 41.3
& 31.5
& 28.5 \\

Gemini-1.5-Pro
& 45.4
& 56.2
& 30.9
& 64.1
& 43.6
& 51.3
& 46.3
& 36.0
& 34.6 \\

Gemini-2.5-Pro
& 52.7
& 48.2
& 35.6
& \textbf{71.3}
& 51.7
& \underline{58.9}
& 42.4
& \underline{46.8}
& 66.7 \\

InternVL2-8B
& 34.6
& 23.1
& 28.7
& 48.2
& 39.8
& 36.7
& 30.7
& 29.9
& 39.6 \\

VILA-1.5-40B
& 31.2
& 22.4
& 24.8
& 48.7
& 22.7
& 40.5
& 25.7
& 31.5
& 32.9 \\

LongVA-7B
& 29.2
& 38.0
& 16.6
& 38.9
& 22.2
& 33.1
& 43.3
& 25.4
& 15.7 \\

LLaVA-NeXT-Video-72B
& 40.9
& 48.9
& 22.8
& 57.4
& 35.3
& 42.4
& 36.7
& 35.0
& 48.6 \\

SAT-LLaVA-Video-7B
& -
& -
& -
& -
& 47.3
& 41.1
& 37.1
& 36.1
& 40.4 \\

RynnEC-7B
& 45.8
& 58.5
& 25.4
& 54.9
& 42.7
& 44.2
& -
& 38.7
& 30.5 \\

CA-MLLM-4B
& 46.8
& 71.3
& 39.3
& 50.4
& 65.9
& 50.7
& 45.5
& 31.4
& 20.2 \\

Qwen3-VL-8B
& 57.9
& -
& -
& -
& -
& -
& -
& -
& - \\

SenseNova-SI
& \textbf{64.8}
& -
& -
& -
& -
& -
& -
& -
& - \\

\midrule
\rowcolor{gray!15}
\multicolumn{10}{l}{\textit{Controlled Qwen2.5 Setting with 297K Training}} \\

VG LLM-8B
& 50.6
& 67.9
& 37.7
& 58.6
& 62.0
& 46.6
& 40.7
& 32.4
& 59.2 \\

\textbf{LieVLM-8B (Ours)}
& 52.4\,{\textcolor{green!50!black}{\tiny +1.7}}
& 69.6\,{\textcolor{green!50!black}{\tiny +1.7}}
& 40.5\,{\textcolor{green!50!black}{\tiny +2.8}}
& 58.7\,{\textcolor{green!50!black}{\tiny +0.1}}
& 64.7\,{\textcolor{green!50!black}{\tiny +2.7}}
& 49.1\,{\textcolor{green!50!black}{\tiny +2.5}}
& 42.9\,{\textcolor{green!50!black}{\tiny +2.2}}
& 33.5\,{\textcolor{green!50!black}{\tiny +1.1}}
& 59.8\,{\textcolor{green!50!black}{\tiny +0.6}} \\

\midrule
\rowcolor{gray!15}
\multicolumn{10}{l}{\textit{Controlled Qwen3 Setting with 3.05M Training}} \\

VG LLM-Qwen3
& 60.6
& \underline{74.4}
& \underline{50.9}
& \underline{64.3}
& \underline{70.3}
& 57.8
& \underline{52.2}
& 45.0
& \underline{69.9} \\

\textbf{LieVLM-Qwen3 (Ours)}
& \underline{62.8}\,{\textcolor{green!50!black}{\tiny +2.2}}
& \textbf{75.4}\,{\textcolor{green!50!black}{\tiny +1.0}}
& \textbf{52.7}\,{\textcolor{green!50!black}{\tiny +1.8}}
& 64.2\,{\textcolor{gray}{\tiny -0.1}}
& \textbf{71.5}\,{\textcolor{green!50!black}{\tiny +1.2}}
& \textbf{60.6}\,{\textcolor{green!50!black}{\tiny +2.8}}
& \textbf{55.5}\,{\textcolor{green!50!black}{\tiny +3.3}}
& \textbf{48.8}\,{\textcolor{green!50!black}{\tiny +3.8}}
& \textbf{73.7}\,{\textcolor{green!50!black}{\tiny +3.8}} \\

\bottomrule
\end{tabular}}
\end{table*}

\paragraph{Multi Seed Reliability.}
The original Qwen2.5 setting is additionally evaluated over three matched random seeds. LieVLM improves all eight categories in every run, increasing the average from $50.62{\pm}0.18$ to $52.35{\pm}0.11$. 
Table~\ref{tab:vsi_multiseed} reports the complete uncertainty, paired gains, and positive seed counts.

\begin{table*}[t]
\centering
\small
\setlength{\tabcolsep}{5pt}
\caption{Multi seed reliability on VSI-Bench. Results are mean $\pm$ standard deviation over three matched seeds, with paired gains computed seed wise.}
\vspace{-2.5mm}
\label{tab:vsi_multiseed}
\resizebox{0.83\textwidth}{!}{
\begin{tabular}{lccccc}
\toprule
\textbf{Metric}
& \textbf{VG LLM-8B}
& \textbf{LieVLM-8B}
& \textbf{Paired Gain}
& \textbf{Positive Seeds}\\
\midrule
Obj. Count
& $67.88{\pm}0.26$
& $\mathbf{69.58{\pm}0.14}$
& $\mathbf{+1.70{\pm}0.12}$
& 3/3\\

Abs. Dist.
& $37.68{\pm}0.24$
& $\mathbf{40.50{\pm}0.11}$
& $\mathbf{+2.82{\pm}0.13}$
& 3/3\\

Obj. Size
& $58.60{\pm}0.06$
& $\mathbf{58.70{\pm}0.09}$
& $\mathbf{+0.10{\pm}0.03}$
& 3/3\\

Room Size
& $61.98{\pm}0.21$
& $\mathbf{64.69{\pm}0.13}$
& $\mathbf{+2.71{\pm}0.09}$
& 3/3\\

Rel. Dist.
& $46.57{\pm}0.22$
& $\mathbf{49.11{\pm}0.10}$
& $\mathbf{+2.54{\pm}0.12}$
& 3/3\\

Rel. Dir.
& $40.69{\pm}0.22$
& $\mathbf{42.90{\pm}0.15}$
& $\mathbf{+2.20{\pm}0.08}$
& 3/3\\

Route Plan
& $32.39{\pm}0.20$
& $\mathbf{33.50{\pm}0.08}$
& $\mathbf{+1.11{\pm}0.12}$
& 3/3\\

Appr. Order
& $59.20{\pm}0.07$
& $\mathbf{59.80{\pm}0.12}$
& $\mathbf{+0.60{\pm}0.05}$
& 3/3\\

\midrule
Average
& $50.62{\pm}0.18$
& $\mathbf{52.35{\pm}0.11}$
& $\mathbf{+1.72{\pm}0.07}$
& 3/3\\
\bottomrule
\end{tabular}}
\end{table*}

\subsection{Robustness to Sparse Views and Geometry Perturbations}
\label{app:robustness}

\paragraph{Sparse Input Views.}
We further evaluate whether the advantage of Lie routing persists when fewer views are available. Frames are uniformly subsampled from the same temporal interval while preserving the scene, query, temporal span, and endpoint views, and VG LLM and LieVLM receive identical frame subsets at every density. As shown in Table~\ref{tab:view_density}, LieVLM remains stronger across all evaluated input densities. The performance gap is preserved as the observations become sparser, with larger gains under the lowest density on Scan2Cap and EmbodiedScan.

\begin{table*}[t]
\centering
\small
\setlength{\tabcolsep}{6pt}
\caption{
Performance under decreasing input view density.
Relative density is measured with respect to the task specific full input setting.
Both models receive identical frame subsets sampled from the same temporal interval.
The full-density results are from the runs used for the sparse-view study, whereas Table~\ref{tab:mechanism_ablation} reports three-seed averages.
}
\label{tab:view_density}
\begin{tabular}{lcccc}
\toprule
\textbf{Task}
& \textbf{Frames}
& \textbf{Relative Density}
& \textbf{VG LLM}
& \textbf{LieVLM} \\
\midrule
\multirow{3}{*}{Scan2Cap C@0.5}
& 32 & 100\% & 80.06 & 82.70 \\
& 16 & 50\%  & 74.31 & 77.08 \\
& 8  & 25\%  & 66.42 & 70.71 \\
\midrule
\multirow{3}{*}{ScanRefer Acc@0.5}
& 32 & 100\% & 50.87 & 57.21 \\
& 16 & 50\%  & 48.34 & 54.85 \\
& 8  & 25\%  & 42.66 & 49.36 \\
\midrule
\multirow{3}{*}{EmbodiedScan 6F F1@0.25}
& 6 & 100\%  & 40.78 & 44.41 \\
& 4 & 66.7\% & 38.05 & 42.41 \\
& 3 & 50\%   & 34.47 & 39.32 \\
\bottomrule
\end{tabular}
\end{table*}

\paragraph{Geometry Perturbations.}
We separately measure the sensitivity of LieVLM to errors in the predicted geometry by perturbing $F^{3d}$ while keeping the visual tokens $F^{2d}$ fixed.
Additive feature noise with $\sigma=0.10$ and token dropout with $p=30\%$ are applied to the trained checkpoints without corruption specific retraining, and each stochastic setting is averaged over three independent realizations. As shown in Table~\ref{tab:geometry_robustness}, LieVLM retains at least $95.5\%$ of clean performance under both perturbations, indicating tolerance to moderate errors in the geometry representation.

\begin{table*}[ht]
\centering
\small
\setlength{\tabcolsep}{5pt}
\caption{
Inference time robustness to geometry perturbations.
Values in brackets report performance retained relative to the clean setting.
The clean results are from the runs used for the perturbation study, whereas the multi-seed detection results are reported in Table~\ref{tab:det_multiseed}.
}
\label{tab:geometry_robustness}
\resizebox{\linewidth}{!}{
\begin{tabular}{lcccc}
\toprule
\textbf{Condition}
& \textbf{Severity}
& \textbf{ScanRefer Acc@0.5}
& \textbf{EmbodiedScan 4F F1@0.25}
& \textbf{EmbodiedScan 6F F1@0.25} \\
\midrule
Clean
& --
& 57.20 [100\%]
& 45.90 [100\%]
& 44.40 [100\%] \\

Geometry Feature Noise
& $\sigma=0.10$
& 55.87 [97.7\%]
& 44.93 [97.9\%]
& 43.31 [97.5\%] \\

Geometry Token Dropout
& $p=30\%$
& 54.76 [95.7\%]
& 43.84 [95.5\%]
& 42.69 [96.1\%] \\
\bottomrule
\end{tabular}}
\end{table*}

\section{More Visualization}
\begin{figure}[ht]
\centering
\includegraphics[width=1.0\textwidth]{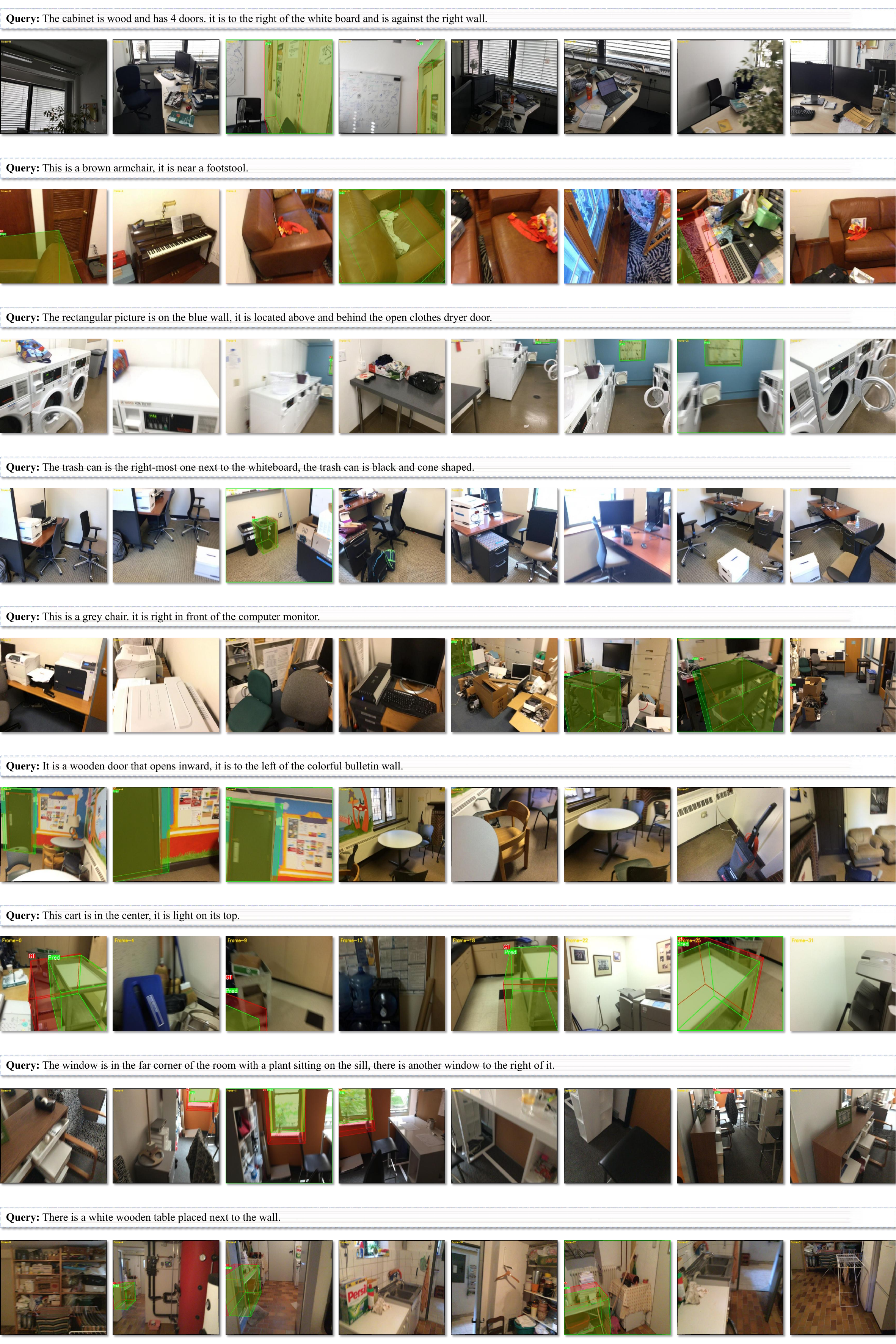}
\vspace{2mm}
\caption{Additional visualization results for 3D visual grounding.}
\label{fig:visulization_vg_appx}
\end{figure}

We provide additional qualitative results for 3D visual grounding in Fig.~\ref{fig:visulization_vg_appx}. The examples cover diverse object categories and spatial expressions, including cabinets, armchairs, pictures, trash cans, chairs, doors, carts, windows, and tables. Across these cases, LieVLM is able to associate the language query with the correct temporal evidence and localize the referred object in 3D space. The projected boxes generally align well with the target regions across different views, showing that the model can handle object attributes, relative positions, and view-dependent appearances. We also observe that challenging cases mainly arise from depth ambiguity, partial visibility, and thin or distant objects, where the 2D projection can still look plausible while the 3D box orientation or depth is less accurate.

\begin{figure}[ht]
\centering
\includegraphics[width=1.0\textwidth]{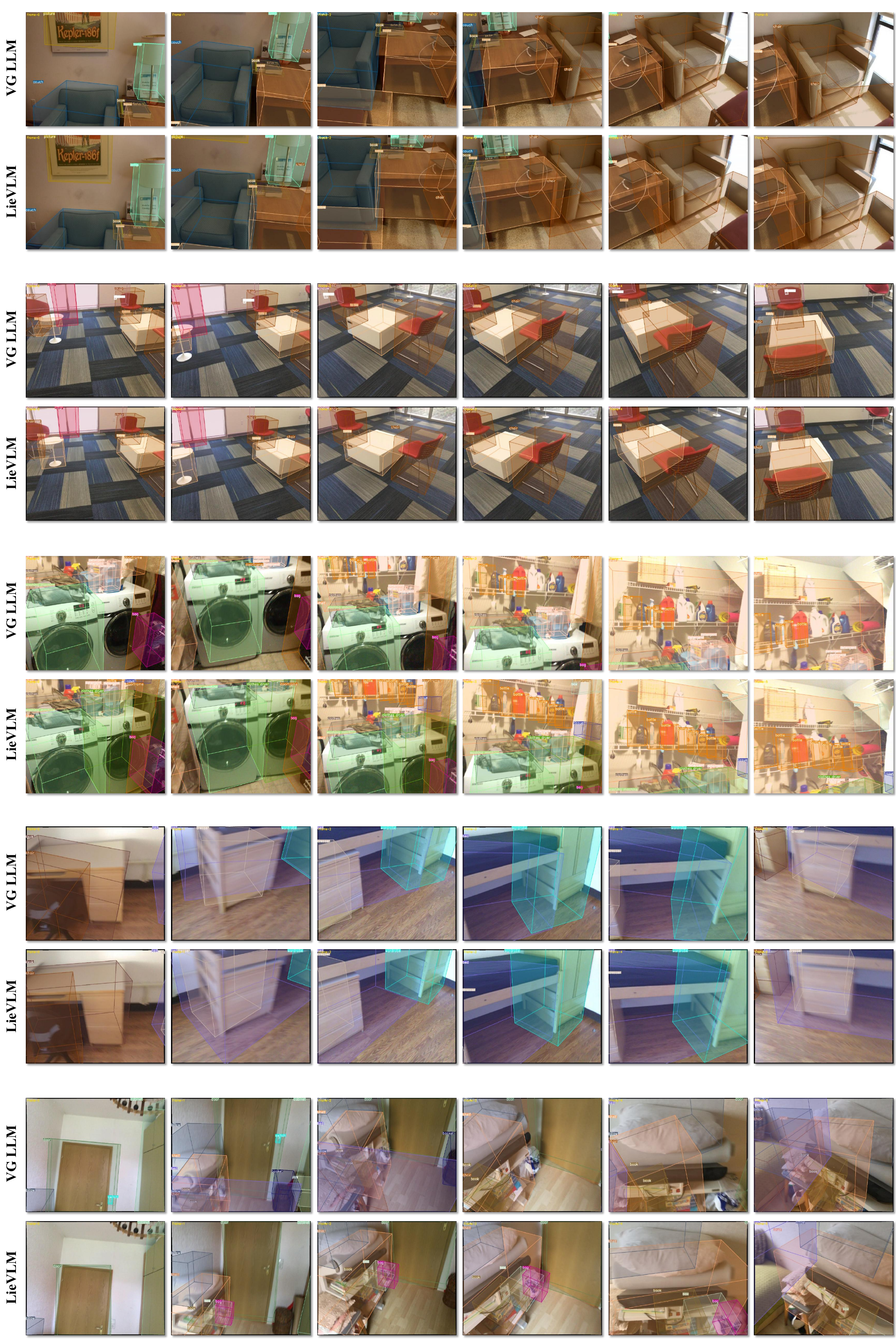}
\vspace{2mm}
\caption{Additional visualization results for 3D video object detection.}
\label{fig:visualization_appx}
\end{figure}

Fig.~\ref{fig:visualization_appx} presents additional results for 3D video object detection. Compared with VG LLM, LieVLM produces more stable 3D boxes across frames and better preserves the spatial orientation of objects. In several examples, VG LLM detects the major objects but predicts boxes with incorrect poses or misses objects under viewpoint changes. In contrast, LieVLM recovers more complete object sets and yields boxes that are better aligned with the visible geometry. These qualitative comparisons further support the role of Lie-Group Routing in reducing view-induced semantic drift and improving cross-frame 3D localization.

\section{Broader Impact}

LieVLM aims to improve the view robustness of 3D multimodal reasoning from RGB videos. This capability can benefit applications that require reliable spatial understanding across changing viewpoints, such as embodied navigation, assistive robotics, indoor mapping, and human-robot interaction. By reducing the dependence on explicit 3D sensors or reconstructed scene inputs, the framework may also make 3D scene understanding more accessible in resource-constrained settings.
At the same time, LieVLM inherits several risks from its underlying MLLM backbone and geometry encoder. The model may still hallucinate objects or spatial relations, especially under poor visibility, severe occlusion, motion blur, or inaccurate geometry estimation. Since its predictions can influence downstream spatial decisions, the model should not be used as the sole source of truth in safety-critical systems without verification, calibration, and human oversight. The base language and vision models may also carry dataset biases, and improved 3D reasoning could be misused for intrusive surveillance or unauthorized spatial analysis. We therefore encourage responsible deployment, careful evaluation across environments and populations, and transparency about failure cases when releasing 3D MLLM systems.


\section{Limitations}
\label{app:limitations}

LieVLM uses the same geometry representation for direct 3D context and latent operator prediction, so reconstruction or alignment errors can affect both paths. The routing gate and Pad\'e normalization constrain the influence of moderate perturbations, but severe geometry degradation can still reduce the reliability of routed corrections. 
The current formulation uses local $\mathrm{SO}(3)$ actions within learned three dimensional channel groups.
The rotation constraint applies to these local action matrices. The learned projections and routing gates do not impose equivariance or composition consistency with respect to physical camera transformations on the complete model.
These actions provide rotation oriented structure but do not explicitly represent translation, non rigid motion, or long range interactions, and the latent groups should not be interpreted as physical coordinates. The routing heads and output projections add computation relative to additive fusion. Current evaluations focus mainly on static indoor scenes, leaving dynamic objects, outdoor environments, and long horizon trajectories for future study.

\section{Usage of Large Language Models}

In this work, Large Language Models (LLMs) are only used for polishing the writing. They are not involved in research ideation, experimental design, data analysis, or the formulation of conclusions. All substantive intellectual contributions are made by the authors.
The authors have checked the content throughout the paper after polishing, and take full responsibility for the current form.


\clearpage
\newpage
\section*{NeurIPS Paper Checklist}

\begin{enumerate}

\item {\bf Claims}
    \item[] Question: Do the main claims made in the abstract and introduction accurately reflect the paper's contributions and scope?
    \item[] Answer: \answerYes{} 
    \item[] Justification: The abstract and introduction covers the theoretical and technical contribution, the developed method and the experimental contribution.

\item {\bf Limitations}
    \item[] Question: Does the paper discuss the limitations of the work performed by the authors?
    \item[] Answer: \answerYes{} 
    \item[] Justification: At the end of the conclusion section, a limitation discussion is provided.

\item {\bf Theory assumptions and proofs}
    \item[] Question: For each theoretical result, does the paper provide the full set of assumptions and a complete (and correct) proof?
    \item[] Answer: \answerYes{} 
    \item[] Justification: Yes, the assumptions are given in the form of Lema in the Prelimaries section. A complete proof is given in the Supplementary material.

    \item {\bf Experimental result reproducibility}
    \item[] Question: Does the paper fully disclose all the information needed to reproduce the main experimental results of the paper to the extent that it affects the main claims and/or conclusions of the paper (regardless of whether the code and data are provided or not)?
    \item[] Answer: \answerYes{} 
    \item[] Justification: Yes, the baseline model, technical details, hyper-parameter and configuration are detailed in the submission for reproducibility.

\item {\bf Open access to data and code}
    \item[] Question: Does the paper provide open access to the data and code, with sufficient instructions to faithfully reproduce the main experimental results, as described in supplemental material?
    \item[] Answer: \answerYes{} 
    \item[] Justification: The datasets this paper uses are publicly available, and the source code is promised to be public once published.

\item {\bf Experimental setting/details}
    \item[] Question: Does the paper specify all the training and test details (e.g., data splits, hyperparameters, how they were chosen, type of optimizer) necessary to understand the results?
    \item[] Answer: \answerYes{} 
    \item[] Justification: The implementation details are given at the beginning of the experimental section.

\item {\bf Experiment statistical significance}
    \item[] Question: Does the paper report error bars suitably and correctly defined or other appropriate information about the statistical significance of the experiments?
    \item[] Answer: \answerNo{} 
    \item[] Justification: Following prior works in this field, the evaluation protocols on the corresponding datasets do NOT require a report of the error bar.

\item {\bf Experiments compute resources}
    \item[] Question: For each experiment, does the paper provide sufficient information on the computer resources (type of compute workers, memory, time of execution) needed to reproduce the experiments?
    \item[] Answer: \answerYes{} 
    \item[] Justification: The hardware, especially the GPU requirement, is listed in the supplementary materials.
    
\item {\bf Code of ethics}
    \item[] Question: Does the research conducted in the paper conform, in every respect, with the NeurIPS Code of Ethics \url{https://neurips.cc/public/EthicsGuidelines}?
    \item[] Answer: \answerYes{} 
    \item[] Justification: This paper focuses on a fundamental task of machine learning and conducts experiments on publicly available datasets.

\item {\bf Broader impacts}
    \item[] Question: Does the paper discuss both potential positive societal impacts and negative societal impacts of the work performed?
    \item[] Answer: \answerYes{} 
    \item[] Justification: The societal impact of this work has been discussed at the end of the conclusion section. We do not envision negative societal impact could be brought by this work.

    
\item {\bf Safeguards}
    \item[] Question: Does the paper describe safeguards that have been put in place for responsible release of data or models that have a high risk for misuse (e.g., pre-trained language models, image generators, or scraped datasets)?
    \item[] Answer: \answerNA{} 
    \item[] Justification:  This work focuses on a fundamental problem in machine learning and conducts experiments on standard datasets. We do not envision such risks.


\item {\bf Licenses for existing assets}
    \item[] Question: Are the creators or original owners of assets (e.g., code, data, models), used in the paper, properly credited and are the license and terms of use explicitly mentioned and properly respected?
    \item[] Answer: \answerYes{} 
    \item[] Justification: Yes, all the assets have been properly cited, with a license to use for academia and no commercial purpose.

\item {\bf New assets}
    \item[] Question: Are new assets introduced in the paper well documented and is the documentation provided alongside the assets?
    \item[] Answer: \answerNA{} 
    \item[] Justification: This paper does not release new assets.

\item {\bf Crowdsourcing and research with human subjects}
    \item[] Question: For crowdsourcing experiments and research with human subjects, does the paper include the full text of instructions given to participants and screenshots, if applicable, as well as details about compensation (if any)? 
    \item[] Answer: \answerNA{} 
    \item[] Justification:  This paper does not involve crowdsourcing nor research with human subjects.

\item {\bf Institutional review board (IRB) approvals or equivalent for research with human subjects}
    \item[] Question: Does the paper describe potential risks incurred by study participants, whether such risks were disclosed to the subjects, and whether Institutional Review Board (IRB) approvals (or an equivalent approval/review based on the requirements of your country or institution) were obtained?
    \item[] Answer: \answerNA{} 
    \item[] Justification: This paper does not involve crowdsourcing nor research with human subjects.

\item {\bf Declaration of LLM usage}
    \item[] Question: Does the paper describe the usage of LLMs if it is an important, original, or non-standard component of the core methods in this research? Note that if the LLM is used only for writing, editing, or formatting purposes and does \emph{not} impact the core methodology, scientific rigor, or originality of the research, declaration is not required.
    \item[] Answer: \answerNA{} 
    \item[] Justification: The core method development in this research of this paper does not involve LLMs as any important, original, or non-standard components.

\end{enumerate}

\end{document}